\documentclass{article}

\usepackage[preprint,nonatbib]{neurips_2024}

\usepackage[T1]{fontenc}    % use 8-bit T1 fonts
\usepackage{hyperref}       % hyperlinks
\usepackage{url}            % simple URL typesetting
\usepackage{booktabs}       % professional-quality tables
\usepackage{amsfonts}       % blackboard math symbols
\usepackage{nicefrac}       % compact symbols for 1/2, etc.
\usepackage{microtype}      % microtypography
\usepackage{xcolor}         % colors
\usepackage{epsfig}
\usepackage{graphicx}
\usepackage{multirow}
\usepackage{makecell}
\usepackage{amsmath}
\usepackage{amssymb}
\usepackage{mathtools}
\usepackage{amsthm}
\usepackage{mathrsfs}
\usepackage{float}
\usepackage{booktabs} 
\usepackage{multicol}
\usepackage{caption}
\usepackage{wrapfig}

\title{Physics-embedded Fourier Neural Network for Partial Differential Equations}

\author{%
 Qingsong Xu\\
 Technical University of Munich \\
 \And
 Nils Thuerey\\
 Technical University of Munich \\
 \And
 Yilei Shi\\
 Technical University of Munich
 \And
 Jonathan Bamber\\
 Technical University of Munich \\
 University of Bristol
 \And
 Chaojun Ouyang\\
 Chinese Academy of Sciences
 \And
 Xiao Xiang Zhu \\
 Technical University of Munich} 
\begin{document}

\maketitle

\begin{abstract}
	We consider solving complex spatiotemporal dynamical systems governed by
	partial differential equations (PDEs) using frequency domain-based discrete
	learning approaches, such as Fourier neural operators. Despite their widespread use for approximating nonlinear PDEs, the  majority of these methods neglect fundamental physical laws and lack interpretability. We address these shortcomings by introducing Physics-embedded Fourier Neural Networks (PeFNN) with flexible and explainable error control. PeFNN is designed to	enforce momentum conservation and yields interpretable nonlinear expressions	by utilizing unique multi-scale momentum-conserving Fourier (MC-Fourier) layers and an element-wise product operation. The
	MC-Fourier layer is by design translation- and rotation-invariant in the
	frequency domain, serving as a plug-and-play module that adheres to the laws of
	momentum conservation. PeFNN establishes a new state-of-the-art in
	solving widely employed spatiotemporal PDEs and generalizes well across
	input resolutions. Further, we demonstrate its outstanding performance for
	challenging real-world applications such as large-scale flood simulations. 
\end{abstract}

\section{Introduction}
\label{Introduction}
The solution of Partial Differential Equations (PDEs)  is crucial in modeling dynamic physical processes, especially in fluid dynamics~\cite{morton2018deep, bonnet2022airfrans}. Traditional fluid dynamics simulators often rely on manually crafted simplifications and demand extensive computational resources~\cite{gal1975use,bates2010simple}.
%, exemplified by the utilization of the Navier-Stokes (NS) equations for fluid flow~\cite{gal1975use} and the shallow water equations (SWE) for flood inundation modeling~\cite{bates2010simple}. 
In recent years, machine learning-based approaches have emerged as highly promising alternatives, demonstrating notable efficacy in addressing these complex problems~\cite{kochkov2021machine,xu2023physics}. 

\textbf{Continuous learning approximation vs. Discrete learning approaches.} Physics-informed neural networks (PINNs)~\cite{raissi2019physics,karniadakis2021physics} employ a continuous learning paradigm, utilizing neural networks (NNs) for the continuous approximation of physical system solutions.  However, PINNs require knowledge of the governing PDEs for optimization, lacking the capability to inherently encode prior physics into the model~\cite{fuks2020limitations}. A more efficient alternative is spatiotemporal discrete learning methods. These methods offer the distinct advantage of hard-encoding physical conditions and properties, as well as incomplete PDE structures into the learning model. Discrete learning methods can be categorized into non-frequency domain approaches, such as convolutional neural networks (CNNs)~\cite{rao2023encoding,long2018pde,gao2021phygeonet} and graph neural networks (GNN)~\cite{brandstetter2021message,pfaff2020learning,horie2022physics}, and frequency domain-based methods, exemplified by Fourier Neural Operators (FNO)~\cite{li2020fourier, tran2022factorized}. Compared to non-frequency domain approaches~\cite{long2018pde,long2019pde,rao2023encoding}, frequency domain-based approaches provide superior advantages in effective feature representation through fast Fourier transform (FFT) and resolution invariance. Thus, the goal of this work is to establish a physics-embedded discrete learning paradigm based on the frequency domain for predicting nonlinear physical systems.

%\textbf{Conservation laws and interpretability should pervade frequency domain-based discrete learning models in the spatiotemporal PDEs.}
\textbf{Conservation laws and interpretability.}
Despite the notable advances in the development of frequency domain-based discrete learning methods, these models are not aware of fundamental physical laws and lack interpretability. A key challenge is the model's heavy reliance on available data, particularly when the governing PDE is unknown~\cite{li2021physics,goswami2023physics}. To address this limitation, an effective strategy involves incorporating the intrinsic preservation of fundamental physical laws, such as momentum conservation, into the architecture. Previous studies~\cite{wang2020incorporating,kunin2020neural} have demonstrated that integrating physical properties enhances interpretability and generalization in NNs. Furthermore, attempts have been made to incorporate physical properties, including antisymmetrical continuous convolutional layers~\cite{prantl2022guaranteed},  equivariant neural operators~\cite{cheng2023equivariant}, and symmetry-enforcing frameworks based on Noether's theorem~\cite{mattheakis2019physical,muller2023exact,liu2023ino}. However, there is an unexplored aspect of frequency domain-based discrete learning models (such as FNOs), where the explicit preservation of conservation laws remains to be thoroughly investigated. Additionally, despite the crucial role of nonlinear activation functions in discrete neural-PDE solvers for approximating nonlinear PDEs, they pose challenges to interpretability. For instance, a stack of multiple Fourier layers in FNO forms a prolonged nested function that is often intractable for human understanding.

\textbf{Physics-embedded Fourier neural network.} In this work, we introduce a Physics-embedded Fourier Neural Network (PeFNN) that enforces interpretable nonlinear expression and momentum conservation by design. Our model has four major properties: (1) \textit{Conservation and Generalization.} Leveraging Noether's theorem, we introduce a novel momentum-conserving Fourier (MC-Fourier) layer. This layer ensures the conservation of linear and angular momentum through translation and rotation invariances, respectively. The intrinsic preservation of these fundamental physical laws facilitates automatic generalization across diverse dynamic systems, ensuring robustness to distributional shifts. 
%Additionally,  the field of fluid dynamics (such as Navier-Stokes equations) exhibits inherent symmetries~\cite{wang2020incorporating}. PeFNN incorporates these symmetries (invariances) to enhance model generalization across different data, theoretically reducing generalization errors. 
(2) \textit{Interpretable Nonlinear Expression.} Our model utilizes an element-wise product operation to achieve an interpretable and expressive approximation of the nonlinear function in spatiotemporal PDE learning. 
Contrasting with the use of nonlinear activation functions in other discrete neural-PDE solvers, such as the autoregressive model~\cite{brandstetter2021message}, convolutional neural operators~\cite{raonic2024convolutional}, FNO~\cite{li2020fourier, tran2022factorized}, and Group equivariant FNO (G-FNO)~\cite{helwig2023group}, our approach provides enhanced interpretability and expressive power for nonlinear systems.
(3) \textit{Explainable and Flexible Error Control.} PeFNN introduces controllable truncation errors per time step by incorporating well-known numerical time discretization methods, such as a forward Euler scheme. Additionally, by employing $n$ multi-scale MC-Fourier layers, PeFNN can represent a polynomial approximation of the nonlinear function up to the $n^{th}$ order, thereby providing a level of error control. This property also offers a means for many existing discrete neural-PDE solvers to demystify their black-box frameworks. 
(4) \textit{Zero-shot Super-resolution.} Leveraging MC-Fourier layers, PeFNN achieves zero-shot super-resolution without the need for additional super-resolution modules, distinguishing it from methods like the physics-encoded recurrent convolutional neural network (PeRCNN)~\cite{rao2023encoding}. Furthermore, PeFNN minimally disrupts the Fourier layer, thereby preserving its capability for super-resolution, in contrast to methods like G-FNO~\cite{helwig2023group}.

\textbf{Contributions.} We present PeFNN, a novel frequency domain-based discrete learning method that utilizes unique multi-scale MC-Fourier layers and the element-wise product operation to capture spatial patterns of the nonlinear system. 
%Additionally, PeFNN performs time marching through the forward Euler scheme.   
The MC-Fourier layer is implemented by incorporating rotation-invariant kernels in the frequency domain, including single or multiple rotation kernels, as a plug-and-play module. This layer not only significantly reduces network parameters but also enhances the network's expressiveness. 
We validate our method by solving widely employed PDEs, including the incompressible Navier-Stokes equations and the shallow water equations. 
Experiments demonstrate that PeFNN achieves state-of-the-art performance compared to current optimal neural-PDE solvers, and can generalize to a higher resolution at test time. 
Furthermore, we establish a flood forecasting benchmark dataset using a widely accepted numerical solution, based on the extraordinary flood events in Pakistan (2022) and Mozambique (2019), to assess the real-world applicability and transferability of various neural-PDE methods.
 Experimental results reveal that our model outperforms the baselines, enabling high-precision, cross-regional, and large-scale flood forecasting. PeFNN is publicly available at~\url{https://github.com/zhu-xlab/PeFNN}.
%\section{Related Work}
%\subsection{Physics-embedded Machine Learning}
%Physics-embedded Machine Learning is achieved by embedding the knowledge of physical equations, physical conditions and properties in the model frameworks or modules~\cite{xu2023physics}. Physical properties usually include conservation, symmetry, and causality of physical systems.
%\subsection{Momentum-conserving Neural Networks}
%\section{Methods}
%We describe the details of our method for enforcing the space-time difference schemes and momentum conservation within a NN model.
\section{Physics-aware spatiotemporal PDE learning}
Given a spatiotemporal dynamical system described by a set of nonlinear, coupled PDEs as,
\begin{equation}
u_t(\mathbf{x}, \mathbf{t})=F\left(\mathbf{x}, \mathbf{t}, \mathbf{u}, \nabla_{\mathbf{x}} u, u \cdot \nabla_{\mathbf{x}} u, \nabla^2 u, \cdots\right),
\label{eq1}
\end{equation}
where $u(\mathbf{x}, \mathbf{t}) \in \mathbb{R}^m$ represents the state variable with $m$ components defined over the spatiotemporal domain $\{(\mathbf{x}, \mathbf{t})\} \in \Omega \times \mathcal{T}$ . Here, $\Omega$ and $\mathcal{T}$ denote the spatial and temporal domain, respectively. $\nabla_{\mathbf{x}}$ is the Nabla operator with respect to the spatial coordinate $\mathbf{x}$, and $F(\cdot)$ is a nonlinear function describing the right-hand side of PDEs. Our objective is to approximate the PDE from its solution samples in a way that we can conduct long-term predictions on the dynamical behavior of the equation for any given initial conditions.

Borrowing the concepts of numerical discretization~\cite{long2018pde,long2019pde, lu2018beyond,rao2023encoding},  the spatiotemporal PDE can be solved based on a forward Euler scheme. That said, the state variable $u$ would be updated by, 
\begin{equation}
u_{t+1}=u_{t}+\widehat{F}\left(u_{t} ; \theta\right) \delta t,
\label{eq2}
\end{equation}
where $\delta t$ is the time spacing; $u_{t+1}$ is the prediction at time $t+1$ and $\widehat{F}$ is the approximated $F$ parameterized by $\theta$ that ensembles a series of operations for computing the $F(\cdot)$ in Eq.~\ref{eq1}.

Significantly, we mainly consider the nonlinear function $\widehat{F}$ in the form of polynomial which is very commonly seen in PDEs~\cite{long2018pde,long2019pde, rao2023encoding}. Specifically, based on the relations between correlation (or convolution) and differentiation, as well as the Taylor expansion (detailed proof is in Appendix~\ref{A.A}), a universal polynomial approximation of $\widehat{F}\left(u_{t}; \theta\right)$ can be achieved via the polynomial combination of solution $u_{t}$. Notably, the universal polynomial approximation achieves nonlinearity of $\widehat{F}$ through the element-wise product operation.   Mathematically,
\begin{equation}
\widehat{F}\left(u_{t}\right)=\sum_{c=1}^{N_c} W_c \cdot\left[\prod_{l=1}^{N_l}\left(W_{l} \star u_{t}\right)\right],
\label{eq3}
\end{equation}
where $N_c$ and $N_l$ are the numbers of channels and parallel Conv layers respectively; $\star$ denotes the correlation operation; $W_{l}$ denotes the weight and bias of the kernel filter of $l$-th layer; $W_c$ is the weight corresponding to $c$-th channel. 
The salient features of physics-aware spatiotemporal PDE learning encompass primarily an enhanced interpretability and generalizability of the nonlinear expressivity compared to the nonlinear activation function in NNs, achieved through the utilization of the element-wise product operation,  a high-precision approximation to $\widehat{F}\left(u_{t}\right)$, and flexible designs for spatial differential modules.

We exploit this fact in the following section by parameterizing $W_{l} \star u_{t}$ directly in Fourier space and using the FFT to efficiently compute Eq.~\ref{eq3}. This leads to a fast architecture that obtains state-of-the-art results for spatiotemporal PDE learning.
%\vspace{-2mm}
\section{Physics-embedded Fourier neural network}
\label{method}
%\vspace{-2mm}
We propose to replace the correlation operation with a discrete Fourier transform (DFT). The forward and reverse DFT are defined as 
\begin{equation}
\begin{aligned}
& \mathscr{F}(x_k)=\sum_{n=0}^{N-1} x_n \cdot e^{-2 i \pi(k / N) n}, \quad k=0, \cdots, N-1 \\
& \mathscr{F}^{-1}(x_n)=\frac{1}{N} \sum_{k=0}^{N-1} x_k \cdot e^{2 i \pi(k / N) n}, \quad k=0, \cdots, N-1,
\end{aligned}
\end{equation}
where $x$ is the discrete Fourier transformed data in the wavelength domain and $N$ means complex numbers $x_0, x_1, \ldots, x_{n-1}$. $i=\sqrt{-1}$ is the imaginary unit. Based on the correlation theorem in Fourier space~\cite{wong2005application}, we find that
\begin{equation}
W_{l} \star u_{t}=\mathscr{F}^{-1}\left(\mathscr{F}^*\left(W_{l}\right) \cdot \mathscr{F}\left(u_{t}\right)\right), 
\end{equation} 
here $\mathscr{F}^*$ is complex conjugate of $\mathscr{F}$. Consequently, correlations can be computed using the FFT. Inspired by FNO~\cite{li2020fourier, tran2022factorized, li2022fourier}, a parametric function $R^{l}_{w}$  is employed to realize the Fourier transform of a periodic function $W_{l}$. $R^{l}_{w}$: $\mathbb{Z}^d \times \mathbb{R}^{d_{r}} \rightarrow \mathbb{R}^{d_r \times d_{r}}$  is directly parameterized as a complex-valued tensor that maps to the values of the appropriate Fourier modes $k$. Here, $d$ represents the dimensionality of discretization, and $d_{r}$ is different dimensions of representation in Fourier space. Subsequently, multiple Fourier layers ($K_{0}$,$K_{1}$,...,$K_{l}$) are defined  to solve the nonlinear function $\widehat{F}$,
\begin{equation}
\begin{aligned}
 \widehat{F}\left(u_{t}\right)=\sum_{c=1}^{N_c} W_c \cdot\prod_{l=1}^{N_l}K_{l}, \quad
K_{l} = \mathscr{F}^{-1}\left(R^{l}_{w} \cdot \mathscr{F}\left(u_{t}\right)\right).
\end{aligned}
\label{eq6}
\end{equation}
In comparison to convolutional layers~\cite{rao2023encoding}, Fourier layers offer advantages in effective feature representation and resolution invariance. However, the exclusive utilization of pure Fourier layers neglects the intrinsic preservation of fundamental physical laws in data.  Consequently, in the next section we develop an improved version, the MC-Fourier layer, which encodes fundamental physical laws to improve learning performance. This layer learns spatiotemporal PDEs while automatically guaranteeing fundamental conservation by encoding symmetries in frequency domain.
% \vspace{-2mm}
\subsection{Symmetries and momentum conservation laws \label{4.1}}
% \vspace{-2mm}
Conservation laws describe the physical properties of the PDEs modeling phenomena.  They are widely used for the study of PDEs such as detecting integrability and linearization, enhancing the accuracy of numerical solution methods. It is well-known that Noether's theorem established a close connection between symmetries and conservation laws for the PDEs~\cite{noether1971invariant, bluman2010applications}.

To elucidate  this connection, consider the nonlinear function $\widehat{F}(u_{t})$, with $u_0(\mathbf{x})$ as the initial condition and $u_{\mathbf{t}}(\mathbf{x})$ as the resulting dynamical response. Firstly, let $L_g$ be a translation on the reference frame, i.e., $L_g[u] = u + g = \tilde{u}$, where $g \in \mathbb{R}^d$ is a translation function, and $d$ is the dimension of spatial domain for the spatiotemporal PDEs. Translational invariance implies that translating the input function $u$ before applying the nonlinear function $\widehat{F}$ yields the same result as applying $\widehat{F}$ directly. As such, the resultant physical model remains invariant across spatial locations, and Noether's theorem~\cite{noether1971invariant} guarantees the conservation of linear momentum. 
Secondly, let $L_r$ be a rotation on the reference frame, which rotates the coordinate $\mathbf{x}$ as well as the input function, i.e., $L_R[u] = Ru = \tilde{u}$ , where $R$ is an orthogonal matrix. Rotational invariance means that rotating the input function $u$ first and then applying the nonlinear function $\widehat{F}$ will lead to the same result as applying $\widehat{F}$ directly. As such, the described physical model remains invariant under rotations against the origin, and Noether's theorem~\cite{noether1971invariant} guarantees the conservation of angular momentum. For comprehensive proof and additional details regarding Noether's theorem, refer to \cite{kara2002basis,arnol2013mathematical}. Thus, leveraging the relationships between symmetries and conservations of physical systems in Noether's theorem, we aim to develop a data-driven form that learns complex physical systems with guaranteed momentum conservations by incorporating invariances into NNs.

In this work, the proposed MC-Fourier layer is designed to handle two types of symmetries, 
%ensuring the fundamental momentum  conservation laws,

1. Translational Invariance. Translating the reference frame by $g \in \mathbb{R}^d$ results in an invariant output, i.e., $\widehat{F}[\tilde{u}](\mathbf{x}+g)=\widehat{F}[u](\mathbf{x})$, where $\tilde{u}=u+g$.

2. Rotational Invariance. Rotating the reference frame results in an invariant output, i.e., for any orthogonal matrix $R \in \mathbb{R}^{d \times d}$, one has $\widehat{F}[\tilde{u}](R\mathbf{x})=\widehat{F}[u](\mathbf{x})$, where $\tilde{u} = Ru$.
% \vspace{-2mm}
\subsection{Momentum-conserving Fourier layer}
% \vspace{-2mm}
To derive our MC-Fourier layers, we embed translation and rotation invariances into the Fourier layer utilizing Noether's theorem in Section~\ref{4.1}. 
Architectures with symmetries have previously been realized using group convolutions~\cite{cohen2016group,helwig2023group}, steerable CNN~\cite{weiler2019general, wang2020incorporating}, and equivariant message passing~\cite{schutt2021equivariant,cheng2023equivariant}. 
However, the majority of methods primarily address symmetry embedded in physical space, neglecting data processing in Fourier space. Simultaneously, these approaches primarily focus on enhancing the generalization of processing input data with symmetry rather than emphasizing physical conservation. \cite{helwig2023group} introduce the embedding of group symmetry in the frequency domain for PDEs and propose the connection of symmetries embedding between the frequency and physical domains. That is, applying a transformation to a function in physical space equally applies the transformation to the Fourier transform of the function. This principle serves as a fundamental support for addressing the challenge of embedding translation and rotation invariances in the Fourier Layer.

For translation invariance, CNNs achieve translation invariance through weight-sharing across spatial locations~\cite{lecun1998gradient}. To establish translation invariance in Fourier layers, we employ the correlation theorem.  For the translation $g \in \mathbb{R}^d$, define the feature map $f: \mathbb{R}^{d \times d} \rightarrow \mathbb{R}^{d_v \times d_v} $, where $d_v$ is the dimension of the latent space, this theorem gives that,
%\vspace{-2mm}
\begin{equation}
\begin{aligned}
[L_g[f] \star W](x)
& = \mathscr{F}^{-1}\left(\mathscr{F}\left(L_g[u]\right) \cdot \mathscr{F}\left(W\right)\right) =\sum_y f(y-g) W(y-x) \\
& =\sum_y f(y) W(y-(x-g))  =\left[f \star W]\right(x) = \mathscr{F}^{-1}\left(\mathscr{F}\left(f\right) \cdot \mathscr{F}\left(W\right)\right).
\label{eq7}
\end{aligned}
\end{equation}
%\vspace{-4mm}

Hence, the Fourier layer is a translation invariant map for ensuring the conservation of linear momentum in the spatiotemporal dynamic system.

Although Fourier layers are invariant to translation, they are not invariant to rotational symmetry. Hence, to establish rotational symmetry, considering any rotation transformation matrix 
$R \in \mathbb{R}^{d \times d}$ and the feature map $f: \mathbb{R}^{d \times d} \rightarrow \mathbb{R}^{d_v \times d_v} $, leveraging the connection of transformations between the frequency and physical domains, our MC-Fourier layer is derived in complete analogy to Eq. \ref{eq7}, expressed as,
\begin{equation}
\begin{aligned}
[L_R[f] \star W](x) &=\sum_y f( R^{-1}y) W(y-x) =\sum_y f(y)\left(W(R(y-x))\right) =\sum_y f(y)\left(L_R[W])\right) \\
&=\mathscr{F}^{-1}\left(\mathscr{F}(f) \cdot\left(L_R[\mathscr{F}(W)]\right)\right) =: \mathscr{F}^{-1}\left(\mathscr{F}(f) \cdot R_{\hat{w}}\right),
%[L_R[f] \star W](x)
%& = \sum_y R^{-1} f(y) W(y-x)  = \sum_y f(y) (R^{-1} W(y-x)) \\
%& = \mathscr{F}^{-1}\left(\mathscr{F}\left(f\right) \cdot (R^{-1} \mathscr{F}\left(W\right))\right)  =: \mathscr{F}^{-1}\left(\mathscr{F}\left(f\right) \cdot R_{\hat{w}}\right),
\label{eq8}
\end{aligned}
\end{equation}
where $R_{\hat{w}}=L_R[\mathscr{F}(W)]$ is the complex-valued function with rotational invariance we aim to learn. $\hat{w}$ accounts for the cyclic shift along the specified dimension. Eq.~\ref{eq8} shows that we can efficiently perform rotation-invariant convolutions in the frequency domain by transforming $\mathscr{F}(W)$ by defining a rotation-invariant kernel function $\hat{w}$.

%\begin{figure}[H]
%	\centering
%	{\includegraphics[width = 0.4\textwidth]{fig22.png}}
%	\vspace{-3mm}
%	\caption{The trainable variables are negated and rotated by the center point. This results in a rotation-invariant kernel. The single rotation kernel (a) exhibits superior feature expressibility, while the multiple rotation kernel (b) entails fewer parameters.}
%	\label{fig:1}
%\end{figure}
\begin{wrapfigure}{r}{4.8cm}
	\centering
	{\includegraphics[width = 0.35\textwidth]{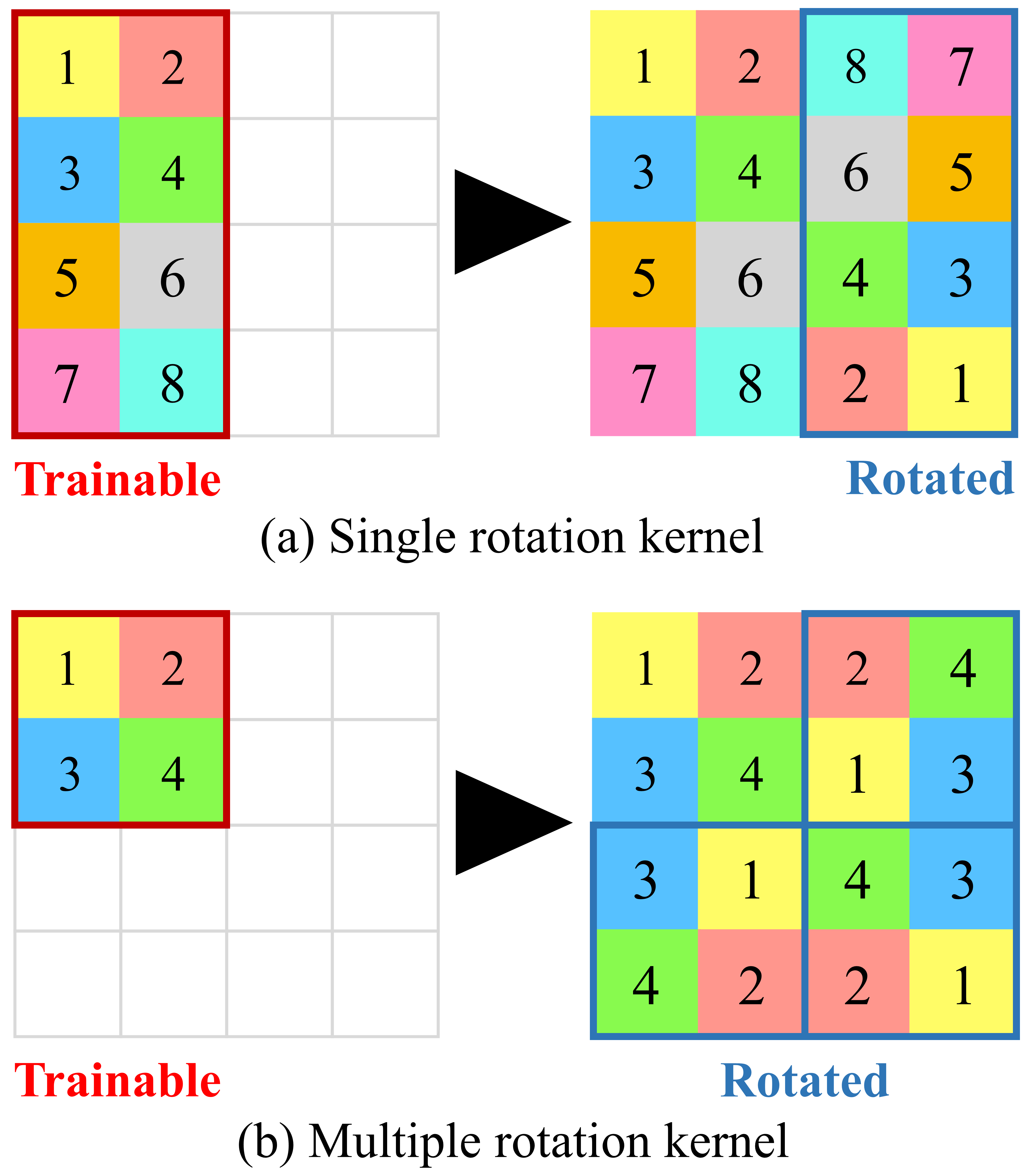}}
	% \vspace{-2mm}
	\caption{The trainable variables are negated and rotated by the center point. This results in a rotation-invariant kernel. The single rotation kernel (a) exhibits superior feature expressibility, while the multiple rotation kernel (b) entails fewer parameters.}
	\label{fig:1}
\end{wrapfigure}
% \vspace{-2mm}
Concretely, to obtain the rotation-invariant property, we define two types of kernel functions: the single rotation kernel (Fig.~\ref{fig:1} (a)) and the multiple rotation kernel (Fig.~\ref{fig:1} (b)). For the single rotation kernel, we halve the learnable kernel parameters with complex-values along a chosen axis and determine the second half by rotation through the center of the kernel. For the multiple rotation kernel, we utilize one-fourth of the kernel as trainable parameters, encompassing both the real and imaginary components. The remaining three-fourths undergo rotations by $s \cdot 90^{\circ}$, $s \in\{1,2,3\}$, relative to its canonical orientation. 
%Both types of rotation-invariant kernels have a centered origin. 
The single rotation kernel facilitates a 50\% reduction in parameters. 
%leading to outstanding feature representation. 
 In contrast, the multiple rotation kernel achieves a 75\% reduction in parameters but exhibits a marginally diminished capability for feature learning. 
Furthermore, to ensure that the kernel function $W$ is real values and that correlation and convolution are equivalent in Eq.~\ref{eq7} and Eq.~\ref{eq8}, the Fourier transform $R_{\hat{w}}$ will be Hermitian. That is, $R_{\hat{w}}=R_{\hat{w}}^*$.
To impose this constraint, it is sufficient to ensure the Hermitian property during the construction of the rotation-invariant kernels, where the two terms corresponding to the upper and lower triangles are conjugates of each other.

To enhance the expressive capacity of our MC-Fourier layer, group convolutions with rotational symmetry ($p4$)~\cite{cohen2016group,helwig2023group} are employed without introducing additional parameters. The group convolutions are applied only along the channel dimension.
Additionally, a residual connection in FNO~\cite{li2020fourier} is utilized. Subsequently, the $l$-th MC-Fourier layer maps the feature map 
$f(u) \in \mathbb{R}^{d_c \times d \times d}$ to $f_{l} \in \mathbb{R}^{d_z \times d_g \times d_v \times d_v}$. Here, $d_c$ is the channel dimension of the physical feature space, and $d \times d$ is the resolution of the input physical space $u$. $d_z$  is the channel dimension of the latent space; $d_g$ is the number of groups in group convolutions ($d_g=4$), and $d_v \times d_v$ is the resolution of the latent space. Our rotation-invariant kernel in the frequency domain is $R_{\hat{w},l} \in \mathbb{R}^{d_c \times d_z \times d_g \times d_v \times d_v}$.
The $l$-th MC-Fourier layer can then be formally expressed as,
\begin{equation}
f_{l}=W_{l} f(u)+\operatorname{MLP}_{l}\left(\mathscr{F}^{-1}\left(\mathscr{F}(f(u)) \cdot R_{\hat{w},l} \right)\right),
\end{equation}
where $W_{l}$ linearly projects the residual connection using a group convolution layer with $1 \times 1$ kernel, and $\operatorname{MLP}_{l}$ is a shallow 2-layer multi-layer perceptron (MLP) with group convolution using a $1 \times 1$ convolution kernel for each layer. 
Notably, the rotation-invariant kernels possess a centered origin. To facilitate proper multiplication with the corresponding modes represented by the matrix $ R_{\hat{w},l}$, the input signal $f(u)$ undergoes a frequency shift to center the zero-frequency component at the origin after applying the FFT $\mathscr{F}$. In addition,  the inverse operation involves reversing the frequency domain shift, followed by the application of the inverse FFT $\mathscr{F}^{-1}$.
\subsection{Architecture and implementation}
% \vspace{-3mm}
\begin{figure}[!htp]
	\centering
	{\includegraphics[width = 0.95\textwidth]{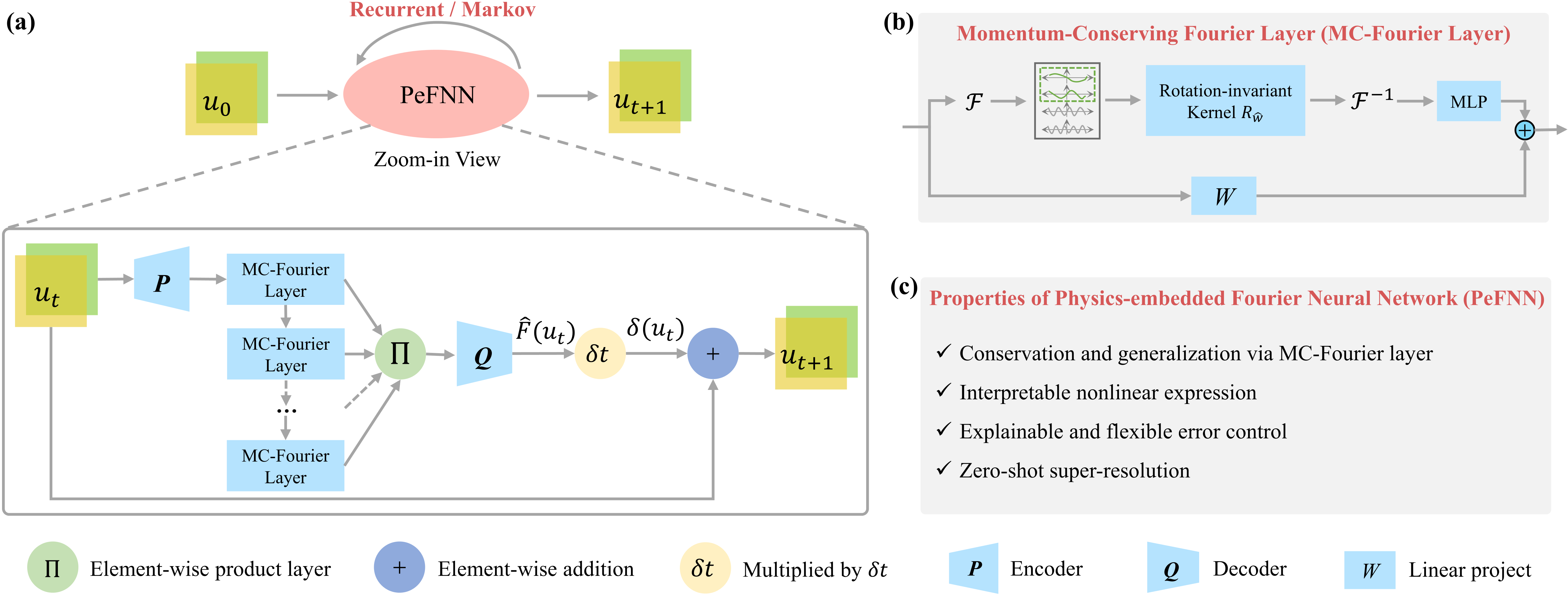}}
	% \vspace{-2mm}
	\caption{Schematic architecture of PeFNN. (a) The network with PeFNN for the Recurrent or Markov training technique; (b) Overview of the Momentum-conserving Fourier layer; (c) properties of PeFNN.
	}
	\label{fig:2}
	% \vspace{-2mm}
\end{figure}
The architecture of PeFNN consists of two major components:  the temporal updates of the state variable $u$ using the forward Euler scheme and the solution of the nonlinear function $\widehat{F}$ through the utilization of MC-Fourier layers and the element-wise product layer. For the solution of the nonlinear function $\widehat{F}$, the core of PeFNN, the state variable $u_t$ from the previous time step goes through multiple MC-Fourier layers. To enhance the feature representation of the state variable $u_t$, an encoder $P$ is employed to lift the 2D input field $u_t$ to a higher-dimensional feature representation $f(u) \in \mathbb{R}^{d_c \times d \times d}$. This encoder, implemented as a linear layer, aims to refine the input. Furthermore, to improve the multi-scale feature representation of the state variable $u_t$, multiple MC-Fourier layers are connected in series. The feature maps generated by these MC-Fourier layers are extracted individually and subsequently fused through the element-wise product layer. A decoder with a MLP layer $Q$ is subsequently used to linearly combine multiple channels into the desired output. The overall architecture, as illustrated in Fig.~\ref{fig:2}, is expressed as,
%\vspace{-2mm}
\begin{equation}
\begin{aligned}
\text{Temporal updates}: \quad & u_{t+1}=u_{t}+\widehat{F}\left(u_{t}\right) \delta t, \\
% \vspace{-2mm}
\text{Solution of }\widehat{F}: \quad & \widehat{F}\left(u_{t}\right)=Q \cdot\prod_{l=1}^{N_l}K_{l},
& K_{l} = f_{l} \circ \cdots \circ f_{2} \circ f_{1} \circ P
\label{eq16}.
\end{aligned}
\end{equation}
%\vspace{-2mm}
PeFNN is designed with rigorous theoretical support, ensuring precision in space-time solutions with controllable error bounds.  Refer to Appendix~\ref{A.B} for a detailed theoretical proof.

\textbf{Training techniques to learn spatiotemporal PDEs.} The utilization of the forward Euler strategy prompts our consideration of two principal training variants for PeFNN: Recurrent and Markov. The efficacy of PeFNN critically relies on employing suitable deep learning techniques. Recurrent training, a widely adopted strategy in spatiotemporal PDE learning, is frequently used in models such as Neural Operator and its variants~\cite{li2020fourier,kovachki2021neural}, PeRCNN~\cite{rao2023encoding}, and others. In this approach, the model predicts the entire rollout during training autoregressively, and the loss is back-propagated through time. However, there are scenarios where Recurrent training utilizing inputs from multiple time steps to the neural network is deemed unnecessary~\cite{tran2022factorized}. In contrast, Markov training involves feeding information solely from the current step, just like a numerical solver. In both strategies, the model is trained to predict only one step into the future, conditioned on the ground truth solutions from previous time steps. In this work, we conduct different experiments and analyses on PeFNN based on these two training strategies.
\section{Experiments}
% \vspace{-2mm}
%In Section~\ref{5.1}, we introduce the datasets encompassing numerical experiments and real flood case simulation for PeFNN. Our experimental settings are detailed in Section~\ref{5.2}, with results and analysis presented in Section~\ref{5.3}.
\subsection{Datasets \label{5.1}}
% \vspace{-2mm}
Our model evaluation encompasses two widely employed PDEs: the incompressible Navier-Stokes (NS) equations and the shallow water equations (SWE). Both equations include momentum conservation. These evaluations extend to both numerical experiments and real-world flood simulation. 
% based on the SWE
%The NS equations and the SWE hold significant relevance in modeling fluid dynamics and hydrodynamic systems. 
% For the real flood case,  during the summer monsoon season of 2022, Pakistan witnessed a catastrophic flood event affecting approximately one-third of the population. This event led to the displacement of around 32 million individuals and resulted in the tragic loss of 1,486 lives~\cite{Bhutto2022}. 
%The economic ramifications of this disaster have been estimated to exceed \$30 billion~\cite{Bhutto2022}. 
%The study area encompasses regions in Pakistan severely affected by the flood, covering a total land area of 85,616.5 square kilometers.

\textbf{2D incompressible Navier-Stokes equations.}
We consider the 2D incompressible NS equations in vorticity form~\cite{li2020fourier}, 
% \vspace{-2mm}
% \begin{equation}
% \begin{gathered}
% \partial_t w(x, t)+u(x, t) \cdot \nabla w(x, t) =\nu \Delta w(x, t)+f(x), \\
% \nabla \cdot u(x, t) =0, \quad
% w(x, 0) =w_0(x),
% \end{gathered}
% \end{equation}
\begin{equation}
\partial_t w(x, t)+u(x, t) \cdot \nabla w(x, t) =\nu \Delta w(x, t)+f(x), 
\quad \nabla \cdot u(x, t) =0, \quad
w(x, 0) =w_0(x),
\end{equation}
where $u(x, t) \in \mathbb{R}^2$ is the velocity, and $w=\nabla \times u$ is the vorticity field we aim to predict. $w_0(x) \in \mathbb{R}$ represents the initial vorticity.  $\nu \in \mathbb{R}_{+}$ denotes the viscosity coefficient, and $f \in \mathbb{R}$ is the forcing function. The solution domain we consider is $x \in(0,1)^2, t \in[0, T]$. We conduct experiments with viscosities $\nu=1 \times 10^{-3}, 1 \times 10^{-4}, 1 \times 10^{-5}$, progressively reducing the final time $T$ as the dynamics become chaotic. The resolution is fixed at $64 \times 64$ for training, validation, and testing. In the super-resolution experiment, we conducted tests with $\nu=1 \times 10^{-4}$, training on a $64 \times 64$ spatial grid and directly testing on a $256 \times 256$ grid. The data generation is provided in Appendix~\ref{B.A}.
%The detailed data generation process is provided in Appendix~\ref{B.A}.

\textbf{2D shallow water equations.} The SWE, derived from the general NS equations~\cite{takamoto2022pdebench}. The equations are expressed  in the form of hyperbolic PDEs,
%\vspace{-2mm}
\begin{equation}
\begin{gathered}
\partial_t h+\nabla h \mathbf{u}=0,  \quad \partial_t h \mathbf{u}+\nabla\left(u^2 h+\frac{1}{2} g h^2\right)=-g h \nabla b,
\end{gathered}
%\vspace{-2mm}
\end{equation}
where $h$ is the depth that we will predict, and $\mathbf{u}=u, v$ is  the velocities in the horizontal and vertical direction. $b$ describes a spatially varying bathymetry. $h \mathbf{u}$ can be interpreted as the directional
momentum components and $g$ is the acceleration due to gravity. The solution domain is  $\Omega=[x,y]=[-2.5,2.5]^2$, $t \in\{1,2, \ldots, 25\}$. The resolution is fixed at $128 \times 128$ for training, validation, and testing. In the super-resolution experiment, we downsample the numerical solution from $128 \times 128$ to $32 \times 32$.
%The initial depth $h(x,y,0)$ is given by,
%\begin{equation}
%h(x,y,0)= \begin{cases}2.0, & \text { for } r<\sqrt{x^2+y^2} \\ 1.0, & \text { for } r \geq \sqrt{x^2+y^2}\end{cases},
%\end{equation}
%where $r$ denotes the radial distance. 
The detailed data generation is provided in Appendix~\ref{B.A}.

\textbf{Real-world flood simulation.} In a flood event, water depth is generally much smaller than the horizontal inundation extent, and the flow hydrodynamics can be mathematically described by the 2D depth-averaged SWE~\cite{de2012improving}. 
%SWEs are described by two conservation laws. 
By neglecting the convective acceleration term, the SWE for flood modeling  can be written as,
%\vspace{-2mm}
\begin{equation}
\begin{gathered}
\frac{\partial h}{\partial t}+\frac{\partial q_x}{\partial x}+\frac{\partial q_y}{\partial y}=R-I, \\
\frac{\partial q_x}{\partial t}+g h \frac{\partial(h+z)}{\partial x}+\frac{g n^2\left|q\right| q_x}{h^{7 / 3}}=0, \quad
\frac{\partial q_y}{\partial t}+g h \frac{\partial(h+z)}{\partial y}+\frac{g n^2\left|q\right| q_y}{h^{7 / 3}}=0,
\end{gathered}
\label{eq44}
%\vspace{-2mm}
\end{equation}
where $h$ is the water height that we will predict, relative to the terrain elevation $z$. 
%$t$ is the time index. $x, y$ are the spatial horizontal coordinates. 
$q = (q_{x}, q_{y})$ is the discharge per unit width. $R$ represents the rainfall rate, and $I$ is the infiltration rate. 
%It is worth noting that the infiltration rate may not be considered when the infiltration has been basically saturated due to continuous rainfall in the study area. 
$n$ is Manning's friction coefficient.  
%The values of Manning’s friction coefficients are based on the range suggested by FLO-2D User’s Manual~\cite{o1993two} and the land cover types of the study area. 
A notable event occurred between August 18 and August 31, 2022, marked by significant increases in flood coverage within the Pakistan study area~\cite{xu2024large, nanditha2023pakistan}. Additionally, a catastrophic flood occurred in Beira, Mozambique, from March 14 to March 20, 2019, due to heavy rainfall~\cite{guo2021mozambique}. Consequently, numerical solutions are implemented for a 14-day period in the Pakistan flood, covering a study area of 85,616.5 km$^{2}$, and a 6-day period in the Mozambique flood, covering a study area of 6,190.9 km$^{2}$. The spatial resolution is fixed at 480m $\times$ 480m, and the temporal resolution is 30 seconds for training, validation, and testing. The flood forecasting benchmark dataset, including study regions, data requirements,  implementation details of numerical methods, and visualization of flood height evolution, is provided in Appendix~\ref{B.B}. 
%We visualize the evolution of the flood height over a 14-day period of rainfall in Fig.~\ref{fig:3}.
%\begin{figure}[!tph]
%	\centering
%	{\includegraphics[width = 0.46\textwidth]{fig4.png}}
%	\vspace{-4mm}
%	\caption{Illustration depicting the evolution of the flood simulation in Pakistan in 2022. The upper curve represents the average depth of the study area, calculated using traditional numerical methods, over a 14-day period of rainfall with a spatial resolution of 480m and a time resolution of 30s. The lower portion illustrates the spatial variation of water depth in the study area over the same 14-day period.
%	}
%	\label{fig:3}
%\end{figure}
% \vspace{-2mm}
\subsection{Experimental settings \label{5.2}}
% \vspace{-2mm}
%\textbf{Task details.} 
For the NS equations, we project the ground truth vorticity field from $t = 10$ to each time step up to $T > 10$.  For SWE, we map the water depth from $t = 1$ to $T=25$. For flood simulation and transfer experiments, we map the flood depth from $t = 1$ to $T=20$ or $T=40$.
%For NS equations, we project the ground truth vorticity field up to $t = 10$ to the field at each time step up to $T > 10$. 
%%Specifically, for $\nu=1 \times 10^{-3}$, we set $T=50$; for $\nu=1 \times 10^{-4}$, we set $T=30$; for $\nu=1 \times 10^{-5}$, we set $T=20$. 
%For SWE, we map the depth of the water at $t = 1$ up to the depth at $T=25$. 
%For flood simulation and transferred experiments, we map the flood depth at $t = 1$ up to the depth at $T=20$ or $T=40$. 
% in Pakistan, we allocate a duration of 14 days with a time step of 30 seconds, resulting in 2,016 samples for $T=20$ and 864 samples for $T=40$. The 6-day samples with a time step of 30 seconds from the Mozambique flood  are used for cross-regional flood forecasting (432 samples for $T=20$ and 864 samples for $T=40$). For each sample, we map the flood depth at $t = 1$ up to the depth at $T=20$ or $T=40$. 
%The 14-day sample is then split into three sets: training (1,440 samples for $T=20$, 720 samples for $T=40$), validation (288 samples for $T=20$, 144 samples for $T=40$), and test (288 samples for $T=20$, 144 samples for $T=40$).
The task details are provided in Appendix~\ref{B.C0}. PeFNN employs the single rotation kernel in MC-Fourier layers without explicit specification. The implementation details of PeFNN are outlined in Appendix~\ref{B.CA}.
We present visualizations of the predicted rollouts for each dataset under both Recurrent and Markov training strategies in Appendix~\ref{D}.
%The 14-day sample is then partitioned into the first 10 days as the training set (with 1440 samples for $T=20$ and 720 samples for $T=40$), the subsequent 2 days as the validation set (with 288 samples for $T=20$ and 144 samples for $T=40$), and the last 2 days as the test set (with 288 samples for $T=20$ and 144 samples for $T=40$).

%\textbf{Implementation Details.} 
%Four MC-Fourier layers are utilized with Fourier modes $k$ = 12 in PeFNN, and the channel dimension of the latent space $d_{z}$ = 10. We use Adam optimizer~\citep{kingma2014adam}  with $\beta_1=0.9, \beta_2=0.999$, and weight decay $10^{-4}$. We use 500 epochs for Recurrent training strategy and 100 epochs for Markov training strategy, with a cosine learning rate scheduler that starts at 0.001 and is decayed to 0. 

%\textbf{Evaluation.} Relative mean square error $L_{\text{RMSE}}$~\cite{li2020fourier, helwig2023group} between the predicted solutions and the computational fluid dynamics  solutions is regarded as an evaluation measure,
%\begin{equation}
%L_{\text{RMSE}}=\frac{1}{n} \sum_{i=1}^{n} \frac{\left\|\hat{y}_i-y_i\right\|_2}{\left\|y_i\right\|_2},
%\end{equation}
%where $\hat{y}_i$ and $y_i$ denote the predicted solution and ground truth of the i-th test PDE. $n$ is the number of test PDEs and $\|\cdot\|_2$ is the $L_2$ norm. For all cases, we run models at least three times with different preset random seeds and average the relative errors. 

\textbf{Benchmarks.} 
State-of-the-art neural-numerical solvers are considered as benchmarks. 
\textbf{U-Net:} A commonly employed architecture for image-to-image regression tasks, available in both 2D (\textbf{U-Net-2D}) and 3D (\textbf{U-Net-3D}) versions
%, involving four blocks with 2D or 3D convolutions and deconvolutions
~\cite{ronneberger2015u}.
% comprises four blocks with 2D convolutions and deconvolutions (\textbf{U-Net-2D}), as well as 3D convolutions and deconvolutions (\textbf{U-Net-3D})~\cite{ronneberger2015u}. 
\textbf{FNO-3D:} 3D FNO with Oneshot training strategy, utilizing direct convolutions in space-time~\cite{li2020fourier}.\textbf{ FNO-2D:} 2D FNO with a Recurrent or Markov training strategy in time~\cite{li2020fourier}. \textbf{PeRCNN:} A convolutional neural network with a Recurrent training strategy in time, designed to preserve the given physics structure~\cite{rao2023encoding}.
%, such as structure or specific terms of the governing PDEs and boundary conditions
\textbf{G-FNO-3D:} 3D group equivariant FNO with the group of translations and $90^{\circ}$ rotations (p4), utilizing Oneshot training strategy in space-time~\cite{helwig2023group}. \textbf{G-FNO-2D:} 2D group equivariant FNO with the group of translations and $90^{\circ}$ rotations (p4), utilizing a Recurrent or Markov training strategy in time~\cite{helwig2023group}. 
 Training details for these benchmarks are available in Appendix~\ref{B.CB}. 
% The relative mean square error, with additional details provided in Appendix~\ref{B.eva}, is regarded as an evaluation measure. Additionally, we compute the momentum loss as $\left|\sum M_{\text {pred}}-\sum M_{\text {ref}}\right|2 / N$ to evaluate each method's effectiveness in conserving momentum. Here, $M{\text {pred}}$ and $M_{\text {ref}}$ represent the predicted and reference momentums, respectively, and $N$ denotes the spatial domain size. 
The relative mean square error serves as an evaluation measure. Additionally, we compute the momentum loss between predicted and true momentum to evaluate each method's effectiveness in conserving momentum. The details of these measures are provided in Appendix~\ref{B.eva}.
For all cases, we run the models three times with different random seeds and average the relative errors. 
%Furthermore, 
%FNO-single rotation kernel and FNO-multiple rotation kernel respectively represent  substituting the Fourier layer in FNO with a single rotation kernel or multiple rotation kernels, to verify the effectiveness of MC-Fourier layer.
\begin{table}[htp!]
	\renewcommand{\arraystretch}{1.02}
    % \vspace{-4mm}
	\caption{Results on NS (fixing resolution $64 \times 64$ for training, validation, and testing). 2D models make rollout predictions with the Recurrent or Markov training strategy in time, while 3D models perform convolutions in space-time. Values in parentheses denote standard deviations.}
	\centering
	\resizebox{1.0\textwidth}{!}
	{\begin{tabular}{c|cc|cccccc}
			\hline
			&                                    &                                       & \multicolumn{2}{c}{$\begin{gathered}
				\nu=1 \times 10^{-3},
				T=50
				\end{gathered}$}                & \multicolumn{2}{c}{$\begin{gathered}
				\nu=1 \times 10^{-4},
				T=30
				\end{gathered}$}                                            & \multicolumn{2}{c}{$\begin{gathered}
				\nu=1 \times 10^{-5},
				T=20
				\end{gathered}$}                                                                                    \\ \cline{4-9} 
			\multirow{-2}{*}{\makecell{Training \\ Strategy}} & \multirow{-2}{*}{Methods} & \multirow{-2}{*}{Parameters} & Valid (\%)            & Test (\%)             & Valid (\%)                        & Test (\%)                           & Valid (\%)                                           & Test (\%)                                               \\ \hline
			& U-Net-3D & 22,587,777 & 2.800(0.364)                                 & 3.075(0.132)                                 & 25.361(0.418)                                 & 25.190(0.462)                                 & 23.401(0.027)                                 & 21.987(0.117)                                 \\
			& FNO-3D   & 11,066,281 & 0.992(0.009)                                 & 0.996(0.027)                                 & 18.729(0.240)                                 & 16.930(0.270)                                 & 19.798(0.147)                                 & {\color[HTML]{080808} 17.226(0.807)}          \\
			\multirow{-3}{*}{Oneshot   (Baselines)} & GFNO-3D  & 9,317,131  & 0.629(0.008)                                 & 0.624(0.016)                                 & 14.440(0.046)                                 & 13.459(0.208)                                 & 15.700(0.222)                                 & 14.895(0.236)                                 \\ \hline
			& PeRCNN   & 566        & 85.645(0.011)                                & 84.711(0.113)                                & 71.805(0.029)                                 & 72.060(0.024)                                 & 71.425(0.000)                                 & 70.297(0.000)                                 \\
			& U-Net    & 7,765,345  & 26.743(4.677)                                & 25.696(1.115)                                & 34.606(1.341)                                 & 31.783(2.467)                                 & 23.988(0.883)                                 & 21.538(0.347)                                 \\
			& FNO-2D   & 928,641    & \underline{0.673}(0.007)    &  \underline{0.696}(0.006)    & 13.527(0.065)                                 & 13.177(0.051)                                 & 14.366(0.138)                                 & 13.383(0.082)                                 \\
			& GFNO-2D  & 853,121    & 0.730(0.044)                                 & 0.746(0.051)                                 & \textbf{11.513}(0.625) & \underline{11.455}(0.642)    &\underline{11.556}(0.728)    & \underline{10.948}(0.817)    \\
			\multirow{-5}{*}{Recurrent}             & PeFNN (Ours)    & 852,221    & \textbf{0.601}(0.024) & \textbf{0.618}(0.022) & \underline{11.694}(0.971)    &  \textbf{11.283}(0.946) & \textbf{11.200}(0.837) & \textbf{10.488}(0.838) \\ \hline
			& U-Net    & 7,762,753  & 7.522(0.866)                                 & 4.277(0.142)                                 & 41.956(6.795)                                 & 18.012(3.095)                                 & 28.278(0.021)                                 & 17.593(1.333)                                 \\
			& FNO-2D   & 928,461    & \underline{0.326}(0.035)    &  \underline{0.327}(0.036)    & 7.457(1.012)                                  & 7.254(0.947)                                  & 9.925(0.116)                                  & 9.395(0.139)                                  \\
			& GFNO-2D  & 853,031    & 0.435(0.116)                                 & 0.420(0.106)                                 &  \underline{4.795}(0.056)     & \underline{4.858}(0.140)     & \underline{7.650}(0.384)     & \underline{7.465}(0.464)     \\
			\multirow{-4}{*}{Markov}                & PeFNN (Ours)    & 851,861    & \textbf{0.190}(0.011)                        & \textbf{0.188}(0.013)                        & \textbf{4.016}(0.094)                         & \textbf{3.796}(0.019)                         & \textbf{7.593}(0.042)                         & \textbf{7.270}(0.072)                         \\ \hline
	\end{tabular}}
	\label{Tab:Table1}
		% \vspace{-4mm}
\end{table}
% \vspace{-3mm}
\subsection{Results and analysis \label{5.3}}
% \vspace{-2mm}
\subsubsection{Incompressible Navier-Stokes equations}
% \vspace{-2mm}
\textbf{Result analysis.} As shown in Table~\ref{Tab:Table1}, we present results on different NS datasets in different training strategies. PeFNN has the best test performance in both Recurrent and  Markov training strategies, effectively reducing baseline errors in 3D models on the NS dataset while utilizing minimal parameters. In a comparative analysis between PeFNN's Recurrent and Markov training strategies, we observed that the error in the Markov strategy is consistently at least 30.68\% lower than that in the Recurrent strategy. This observation aligns with the findings~\cite{tran2022factorized}, indicating that the Markov strategy enhances results in comparison to Recurrent training. 
For the dataset with $\nu=1 \times 10^{-3}, T=50$, in the Markov or Recurrent training strategy, G-FNO-2D shows slightly lower performance than FNO-2D. In contrast, our PeFNN outperforms FNO-2D, highlighting its effective long-term prediction capabilities.
%the performance of G-FNO-2D is marginally lower than that of FNO-2D, while our PeFNN exhibits superior performance compared to FNO-2D. This demonstrates the effective long-term prediction capabilities of our PeFNN. 
In the scenario with $\nu=1 \times 10^{-4}, T=30$ under the Recurrent training strategy, PeFNN performs slightly worse than G-FNO-2D on the validation set. However, on the test set, PeFNN outperforms G-FNO-2D, providing substantial evidence for its enhanced generalizability. Additionally, a comparative analysis with non-frequency domain models (PeRCNN and U-Net) highlights the efficacy of Fourier layers in PeFNN. Notably, PeFNN adheres to an important physical law, enhancing confidence and credibility in its solutions.
%\vspace{-4mm}
\begin{figure}[!htp]
	% \vspace{-2mm}
	\centering
	{\includegraphics[width = 0.95\textwidth]{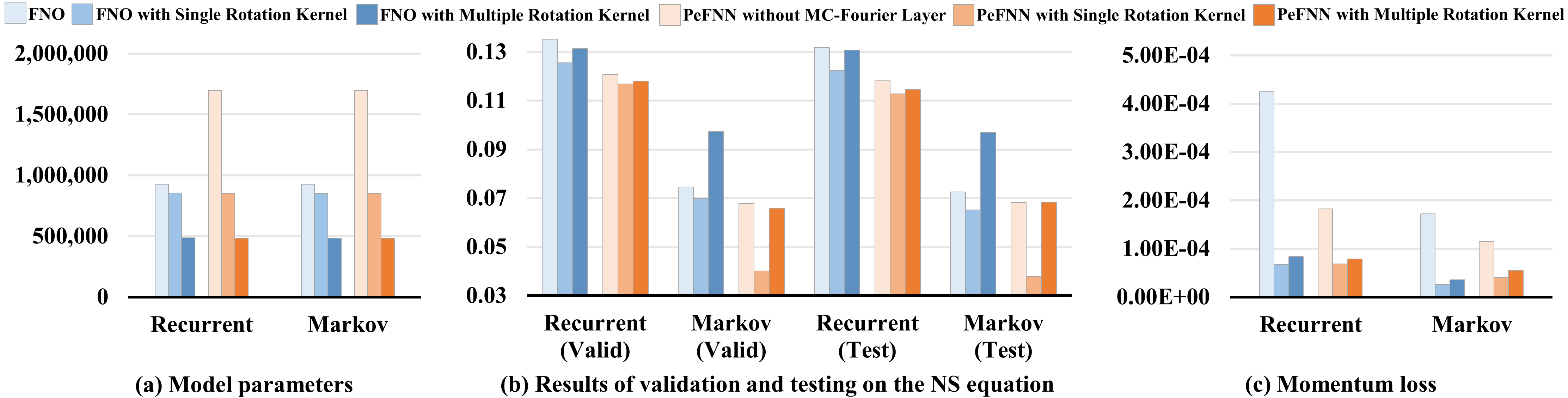}}
	% \vspace{-2mm}
	\caption{Ablation study of MC-Fourier layers on the NS equation with $\nu=1 \times 10^{-4}, T=30$.
	}
	\label{fig:33}
	% \vspace{-4mm}
\end{figure}

\textbf{Ablation study.} 
In Fig.~\ref{fig:33}, it is evident that the inclusion of the MC-Fourier layer with a single rotation kernel in both FNO-2D and PeFNN, under various training strategies, has notably improved performance.  Further comparison between models employing multiple and single rotation kernels reveals a consistent trend across different training strategies (Fig.~\ref{fig:33} (a) and (b)). That is, the multiple rotation kernel exhibits the ability to reduce the number of parameters of models with the single rotation kernel by nearly half. However, this reduction in parameters is accompanied by a decrease in the model's expressive capacity, leading to a decline in accuracy. On the other hand, the single rotation kernel demonstrates superior expressive capabilities, consequently achieving the highest accuracy among models without the MC-Fourier layer and those incorporating single and multiple rotation kernels. Furthermore, when comparing PeFNN without the MC-Fourier layer to FNO-2D without the MC-Fourier layer, it is evident that PeFNN outperforms FNO-2D regardless of the training strategy. Notably, the high number of model parameters in PeFNN without the MC-Fourier layer, attributed to the presence of group convolution, can be significantly reduced by rotation-invariant kernels. Notably,
we calculate the momentum loss (Fig.~\ref{fig:33} (c)) under the assumption of unit mass. It is evident that models incorporating both  single and multiple rotation kernels demonstrate a significant reduction in momentum loss, effectively minimizing it to near-zero levels compared to models without the MC-Fourier layer. Additionally, despite the accuracy of models utilizing the multiple rotation kernel being comparable to or inferior to those lacking the MC-Fourier layer, their momentum loss is notably lower. This observation  provides further evidence of the proficient momentum conservation capabilities of the MC-Fourier layer.
%Additionally, a comparative analysis of FNO-2D and PeFNN under two training strategies reveals that the test performance of the model with multiple rotation kernel in the recurrent training slightly outperforms the baseline without the MC-Fourier layer. This corroborates the efficacy of the multiple rotation kernel. Furthermore, when comparing PeFNN without the MC-Fourier Layer to FNO-2D without the MC-Fourier Layer, it is evident that PeFNN's performance outperforms FNO-2D, regardless of the training strategy. Notably, the high number of model parameters in PeFNN without the MC-Fourier Layer, attributed to the presence of group convolution, can be significantly reduced by rotation-invariant kernels.
%\vspace{-4mm}
\begin{wraptable}{r}{6cm}
	% \vspace{-4mm}
\renewcommand{\arraystretch}{1.1}
	\caption{Super-resolution (SR) results on the NS equation. }
%		We train on a $64 \times 64$ spatial grid and directly test on a $256 \times 256$ grid.
% \vspace{-2mm}
	\centering
	\resizebox{0.45\textwidth}{!}
	{
		\begin{tabular}{c|cc|cc}
			\hline
			\multirow{2}{*}{Methods} & \multicolumn{2}{c|}{Markov Training} & \multicolumn{2}{c}{Recurrent Training} \\ \cline{2-5} 
			& Test (\%)              & SR Test  (\%)        & Test   (\%)            & SR Test (\%)          \\ \hline
			U-Net-2D                    & 18.012(3.095)           & 109.433(10.575)          & 31.783(2.467)            & 88.903(0.414)           \\
			FNO-2D                   & 7.254(0.947)           &  \underline{43.278}(0.201)    & 13.177(0.051)            & 42.768(0.147)           \\
			G-FNO-2D                  & \underline{4.858}(0.140)     & 43.326(0.057)          & \underline{11.455}(0.642)      &  \underline{42.630}(0.078)     \\
			PeFNN                    & \textbf{3.796}(0.019)  & \textbf{42.949}(0.085) & \textbf{11.283}(0.946)   & \textbf{41.516}(0.515)  \\ \hline
	\end{tabular}}
	\label{Tab:Table3}
	% \vspace{-4mm}
\end{wraptable}
\textbf{Super-resolution.} Table~\ref{Tab:Table3} examines the super-resolution capabilities of PeFNN on the NS equation with $\nu=1 \times 10^{-4}$, $T=30$.  
PeFNN achieves the lowest super-resolution error under both training strategies. 
% However, in comparison with FNO-2D,  the super-resolution performance of G-FNO-2D is diminished under the Markov training. 
Additionally, while Markov training significantly improves test performance, it diminishes super-resolution test performance compared to Recurrent training. Furthermore, Appendix~\ref{C.A1} demonstrates PeFNN's excellent robustness and generalization across input data with varying noise levels.
% \vspace{-3mm}
\subsubsection{Shallow water equations}
% \vspace{-3mm}
\begin{wraptable}{r}{6cm}
	\renewcommand{\arraystretch}{1.1}
	% \vspace{-4mm}
	\caption{Results on Shallow Water Equations (fixing resolution $128 \times 128$).}
	\centering
	\resizebox{0.45\textwidth}{!}
	{
		\begin{tabular}{c|cc|cc}
			\hline
			&                                    &                                       &                                       &                                      \\
			\multirow{-2}{*}{Training   Strategy} & \multirow{-2}{*}{Methods} & \multirow{-2}{*}{Parameters} & \multirow{-2}{*}{Valid (\%)} & \multirow{-2}{*}{Test (\%)}      \\ \hline
			& U-Net-3D                           & 22,580,001                            &  0.362(0.032)        &  0.377(0.033)       \\
			& FNO-3D                             & 11,066,101                            &  0.258(0.011)        & 0.278(0.010)       \\
			\multirow{-3}{*}{Oneshot}                      & G-FNO-3D                           & 9,317,041                             & 0.220(0.006)       & 0.243(0.007)       \\ \hline
			& U-Net-2D                           & 7,762,753                             &  0.508(0.079)        & 0.462(0.065)       \\
			& FNO-2D                             & 928,461                               & \underline{0.238}(0.030)  &  \underline{0.261}(0.037) \\
			& G-FNO-2D                           & 853,031                               &  0.348(0.010)        &0.365(0.009)       \\
			\multirow{-4}{*}{Recurrent}                    & PeFNN                              & 851,861                               & \textbf{0.217}(0.004)                      & \textbf{0.240}(0.008)                     \\ \hline
			& U-Net-2D                           & 7,762,753                             & 1.886(0.526)        &  1.847(0.496)       \\
			& FNO-2D                             & 928,461                               & 0.478(0.011)        & 0.390(0.017)       \\
			& G-FNO-2D                           & 853,031                               & \textbf{0.297}(0.025)                      & \textbf{0.306}(0.042)                     \\
			\multirow{-4}{*}{Markov}                       & PeFNN                              & 851,861                               &  \underline{0.328}(0.006)                         &  \underline{0.352}(0.010)                        \\ \hline
	\end{tabular}}
	% \vspace{-4mm}
	\label{Tab:Table4}
\end{wraptable}
In Table~\ref{Tab:Table4}, we present results on the SWE dataset for different training strategies.
In the Recurrent training strategy, PeFNN exhibits the lowest error among all 2D and 3D models, with the fewest model parameters. 
Its error is 34.25\% lower than the second-best benchmark (FNO-2D with Recurrent training). Conversely, in the Markov training strategy, G-FNO-2D achieves peak performance, with PeFNN exhibiting a slightly higher error than G-FNO-2D.
PeFNN's performance in the Markov training strategy is marginally lower than in the Recurrent training strategy. Improvements in PeFNN's performance under the Markov training strategy could be achieved by increasing the training samples of the SWE dataset. Super-resolution results for SWE are presented in Appendix~\ref{C.A}.
\subsubsection{Real-world flood simulation}
% \vspace{-3mm}
\begin{figure}[!htp]
	\centering
	% \vspace{-0.3mm}
	{\includegraphics[width = 0.95\textwidth]{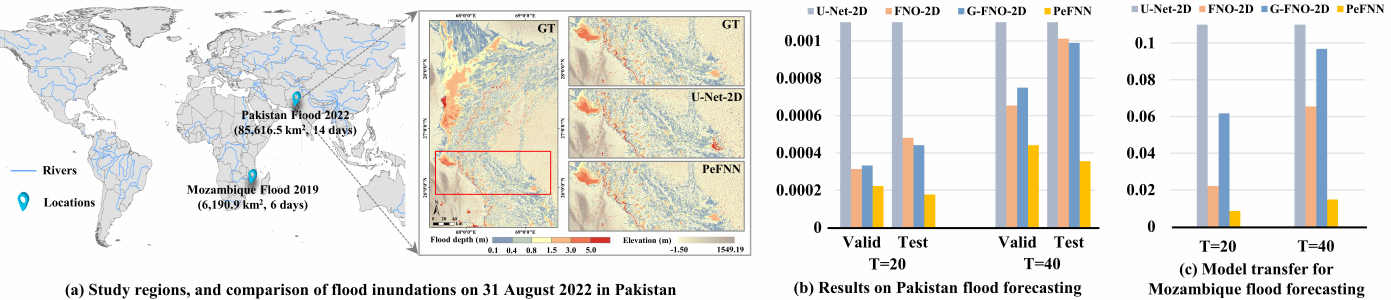}}
		% \vspace{-2mm}
	\caption{Results and transferable results on flood simulation, all trained with the Markov strategy. 
%		Rollouts are of length $T=20$ and $T=40$, using true water depth values at time $t=0$ as initial input. 
}
	\label{fig:5}
		% \vspace{-4mm}
\end{figure}
Fig.~\ref{fig:5} presents the flood simulation results and transferred results for predicting over 20 and 40 time steps, using true water depth values at time $t=0$ as the initial input. 
%The aim is to assess model performance of flood forecasting at different time scales.  
Figure~\ref{fig:5} (b) shows that PeFNN exhibits state-of-the-art performance in large-scale Pakistan flood simulation. Notably,  PeFNN demonstrates its superiority by maintaining stable and high-precision flood predictions across various time scales.   For instance, 
in Pakistan flood prediction at $T=40$, PeFNN achieves a remarkable test error reduction by 64.02\% compared to the best benchmark (G-FNO-2D). Additionally, the right part (red box) of Fig.~\ref{fig:5} (a) demonstrates the accuracy of PeFNN on the final flood inundation with $T=20$.
Furthermore, Fig.~\ref{fig:5} (c) presents the results of directly transferring the model trained on Pakistan flood to predict Mozambique floods. PeFNN achieves the best and most stable cross-regional flood prediction results across various time scales, with a notable test error reduction by over 84.69\% compared to G-FNO-2D. 
Appendix~\ref{C.B} presents consecutive flood prediction results for two days in Pakistan and six days in Mozambique. 
These results demonstrate that PeFNN maintains high-precision, cross-regional, large-scale flood forecasting across different time scales.
% \vspace{-4.5mm}
\section{Conclusion}
\label{Con}
% \vspace{-3mm}
This work introduces PeFNN, a discrete learning method in the frequency domain. By enforcing interpretable nonlinear expression and momentum conservations, PeFNN aims to demystify black-box frameworks of NNs and enhance confidence in solving dynamical systems. Extensive experiments demonstrate PeFNN's state-of-the-art accuracy in solving PDEs and strong generalization across input resolutions. Notably, experiments on the flood forecasting dataset show that PeFNN maintains high-precision, cross-regional, and large-scale flood forecasting.  PeFNN shows potential for rapid global flood forecasting through its strong zero-shot super-resolution and generalization. 
However,  
%the applicability of PeFNN's momentum conservations needs thorough verification across a broader range of PDEs. Additionally,
PeFNN has not explored other physical conservations, such as mass and energy, nor has it been extensively tested with non-fluid PDEs, which will be the focus of future research and development.
%These limitations will be the focus of future improvement and exploration.

%\section*{References}

\clearpage
\newpage
\bibliographystyle{nips}
\bibliography{example_paper}

%%%%%%%%%%%%%%%%%%%%%%%%%%%%%%%%%%%%%%%%%%%%%%%%%%%%%%%%%%%%
\newpage
\appendix

\section{Theoretical proof}
\subsection{Universal polynomial approximation of $\widehat{F}$} 
\label{A.A}
In the proposed physics-aware spatiotemporal PDE learning, $\widehat{F}\left(u_{t}\right)$ acts as a universal polynomial approximator to unknown nonlinear functions. $\widehat{F}$ achieves enhanced interpretability and generalizability of the nonlinear expressivity through the utilization of the element-wise product operation. To support this claim, we propose the following theoretical proof.

Let us firstly denote the set of system state $\mathbf{u}$ and its derivative terms, consisting of $m$ elements in total, as $\theta = \left[\mathbf{x}, \mathbf{u}, \nabla_{\mathbf{x}} u, u \cdot \nabla_{\mathbf{x}} u, \nabla^2 u, \cdots\right]^{T} \in \mathbb{R}^m $. The dynamical system in Eq.~\ref{eq2} can then be represented by $u_{t+1}=u_{t}+F(\theta) \delta t$. Based on the multivariate Taylor's theorem, for any small positive number $\epsilon$, there is a real-valued polynomial function $\mathcal{P}$ that can approximate $F(\theta)$, 
%(reproduced from ~\cite{rao2023encoding}),
\begin{equation}
|F(\theta)-\mathcal{P}(\theta)|<\epsilon,
\label{eq10}
\end{equation}
where $\mathcal{P}(\theta)$ can be expressed as,
\begin{equation}
\begin{aligned}
& \mathcal{P}(\theta)=\sum_{n_1=0}^n \sum_{n_2=0}^n \cdots \sum_{n_m=0}^n N(\theta), \\
N(\theta) &=c_{n_1} c_{n_2} \cdots c_{n_m}\left(\theta_1-\bar{b}_1\right)^{n_1}\left(\theta_2-\bar{b}_2\right)^{n_2} \cdots\left(\theta_m-\bar{b}_m\right)^{n_m}, \\
&=(c\left(\theta_1-\bar{b}_1\right)^{n^{\prime}})(c\left(\theta_2-\bar{b}_2\right)^{n^{\prime}}) \cdots (c\left(\theta_m-\bar{b}_m\right)^{n^{\prime}}), \quad n_1 = n_2 = \cdots = n_m = n^{\prime},
\end{aligned}
\label{eq11}
\end{equation}
where $c$ denotes the coefficients. $\bar{b}$ is the biases, and $n$ represents the maximum polynomial order.

For real numbers $\theta$, $\bar{b}$ and $c$, and an integer $n^{\prime}$, there exists a real-valued vector $\alpha \in \mathbb{R}^{n+1}$, a real number $\tilde{b}$ and an integer $n^{\prime} \leqslant n$ such that $c(\theta-b)^{n^{\prime}}=\prod_{i=0}^n\left(\alpha_i \theta-\tilde{b}\right)$ if $\|\boldsymbol{\alpha}\|_0=n^{\prime}$, where $\|\cdot\|_0$ denotes the $\ell_0$ norm of a vector. Thus, $N(\theta)$ can be re-written as,
\begin{equation}
\begin{aligned}
N(\theta)&=\prod_{i=0}^n\left[\left(\alpha_{i 1} \theta_1-\tilde{b}_1\right)\left(\alpha_{i 2} \theta_2-\tilde{b}_2\right) \cdots\left(\alpha_{i m} \theta_m-\tilde{b}_m\right)\right], \\
&=\prod_{i=1}^{(n+1) m}\left[\beta_i \hat{\theta}_i+\hat{b}_i\right],\\
\boldsymbol{\beta} & \in \mathbb{R}^{(n+1) m} = [\beta_0, \beta_1, \beta_2, \cdots, \beta_{(n+1)m}]^{T} \\
&= [\alpha_{0 1}, \alpha_{0 2}, \ldots, \alpha_{0 m}, \ldots, \alpha_{k 1}, \alpha_{k 2}, \ldots, \alpha_{k m}, \ldots, \alpha_{n 1}, \alpha_{n 2}, \ldots, \alpha_{n m}]^{T},\\
\hat{\boldsymbol{\theta}}:&=\boldsymbol{\ell} \otimes \boldsymbol{\theta} \in \mathbb{R}^{(n+1) m},\\
\boldsymbol{\hat{b}}:&=\hat{\boldsymbol{\ell}} \otimes (-\boldsymbol{\tilde{b}}) \in \mathbb{R}^{(n+1) m},\\
s.t. & \quad \left\|\boldsymbol{\alpha}_k\right\|_0=n_k, \quad  \boldsymbol{\alpha}_k=\left[\alpha_{0 k}, \alpha_{1 k}, \ldots, \alpha_{n k}\right]^{\mathrm{T}} \in \mathbb{R}^{n+1}, 
\end{aligned}
\label{eq12}
\end{equation}
where $\otimes$ denotes the Kronecker product. $\boldsymbol{\ell} \in \mathbb{R}^{n+1}$ and $\hat{\boldsymbol{\ell}} \in \mathbb{R}^{n+1}$  are  column vectors with all elements equal to 1. 

By combining Eq.~\ref{eq11} and Eq.~\ref{eq12}, we obtain,
\begin{equation}
\begin{aligned}
\mathcal{P}(\theta)&=\sum_{n_1=0}^n \sum_{n_2=0}^n \cdots \sum_{n_m=0}^n N(\theta) \\
&=\sum_{n_1=0}^n \sum_{n_2=0}^n \cdots \sum_{n_m=0}^n \prod_{i=1}^{(n+1) m}\left[\beta_i \hat{\theta}_i+\hat{b}_i\right] \\
&=\sum_{c=1}^{N_c} W_c \cdot\left[\prod_{l=1}^{N_{l}}\left(\eta_{l c} E_{l c}+b_l\right)\right], \quad N_{c} \geqslant(n+1)^s, \quad N_{l} \geqslant(n+1) s,
\end{aligned}
\label{eq14}
\end{equation}
where $\mathbf{E}:=\tilde{\ell} \otimes \hat{\boldsymbol{\theta}} \in \mathbb{R}^{N_{c} \times N_{l}}$ is the Kronecker transformation of $\hat{\boldsymbol{\theta}}$. $\mathbf{W} \in \mathbb{R}^{N_{c}}$ and $\boldsymbol{\eta} \in \mathbb{R}^{N_{c} \times N_{l}}$  denote some properly defined real-valued coefficients; $\mathbf{b} \in \mathbb{R}^{N_{l}}$ is the bias vector.

In accordance with the established relationship between correlation and differentiation, as demonstrated in \cite{long2018pde,long2019pde,rao2023encoding}, it follows that a trainable correlational operator can approximate any differential operator with a prescribed order of accuracy. Thus, the term $\left(\eta_{l c} E_{l c}+b_l\right)$ in Eq.~\ref{eq14} can be approximated by a series of correlational operators. That is,
\begin{equation}
\mathcal{P}(\theta)=\sum_{c=1}^{N_c} W_c \cdot\left[\prod_{l=1}^{N_l}\left(W_l \star u\right)\right].
\label{eq15}
\end{equation}
Substituting Eq.~\ref{eq15} into Eq.~\ref{eq10} can thus obtain a universal polynomial approximation of $\widehat{F}$.

\subsection{Error bounds for PeFNN}
\label{A.B}
\textbf{Error bounds of the forward Euler scheme.} Local truncation error of the forward Euler scheme can be obtained based on the Taylor expansion,
\begin{equation}
u_{t+1}=u_{t}+\widehat{F}\left(u_{t}\right) \delta t+\mathcal{O}\left(\delta t^2\right),
\end{equation}
where $\mathcal{O}\left(\delta t^2\right)$ denotes the truncation error. We can see the truncation error of the forward Euler computation converges to zero as $\delta t$ decreases. Furthermore, to enhance the accuracy of the time update for state variables, incorporating higher-order time discrete methods in numerical calculations, such as the third-order Runge-Kutta  method~\cite{butcher1996history}, can be utilized in PeFNN.

\textbf{Error bounds of the nonlinear function $\widehat{F}$.} Through a comparison of Eq.~\ref{eq14} and Eq.~\ref{eq15}, we can find that the error of the nonlinear function is mainly caused by the difference between correction $(W_l \star u)$ and differential operations $\left(\eta_{l c} E_{l c}+b_l\right)$.  By replicating the relationship between differential operation and correlational operator~\cite{long2019pde}, we can derive error bounds for the nonlinear function,
\begin{equation}
\begin{aligned}
W_l \star u = & \sum_{k=-\frac{N-1}{2}}^{\frac{N-1}{2}} W_l\left[k\right] u\left(x+k \delta x\right) \\
= & \sum_{k=-\frac{N-1}{2}}^{\frac{N-1}{2}} W_l\left[k\right] \sum_{i=0}^{N-1} \frac{\partial^{i} u}{\partial^i x}|_{x} \frac{k_1^i}{i !} \delta x^i+\mathcal{O}\left(|\delta x|^{N-1}\right) \\
= & \left.\sum_{i=0}^{N-1} M_i \frac{k_1^i}{i !} \delta x^i \cdot \frac{\partial^{i} u}{\partial^i x}\right|_{x}+\mathcal{O}\left(|\delta x|^{N-1}\right) \\
:= & \left(\eta_{l c} E_{l c}+b_l\right)+\mathcal{O}\left(|\delta x|^{N-1}\right).
\end{aligned}
\label{eq17}
\end{equation}
Here, $N$ is the size of the filter $W_l$.
From Eq.~\ref{eq17} we can see that the difference between correction and differential operations is $\mathcal{O}\left(|\delta x|^{N-1}\right)$. Substituting Eq.~\ref{eq17} into Eq.~\ref{eq15},  the error bound of the nonlinear function is $\mathcal{O}\left(|\delta x|^{(N-1)N_l}\right)$.

\section{Data generation and training  details}
\label{B}
\subsection{Data generation for numerical experiments}
\label{B.A}
For  NS equations, we use the data directly from~\cite{li2020fourier}, with 1,000 training trajectories, 100 validation trajectories, and 100 test trajectories. For each trajectory, the boundary conditions are periodic and the initial conditions $w_0(x)$ are sampled from a Gaussian random field.

For SWE, the initial depth $h(x,y,0)$ is given by,
\begin{equation}
h(x,y,0)= \begin{cases}2.0, & \text { for } r<\sqrt{x^2+y^2} \\ 1.0, & \text { for } r \geq \sqrt{x^2+y^2}\end{cases},
\end{equation}
where $r$ denotes the radial distance of the dam, and $r$ is sampled uniformly from $(0.3,0.7)$.
We use numerical data~\cite{takamoto2022pdebench} generated using the finite volume method. We split 1,000 trajectories into 800 training trajectories, 100 validation trajectories, and 100 test trajectories. 
%For each trajectory, the radius of the dam $r$ is sampled uniformly from $(0.3,0.7)$.
\subsection{Flood forecasting benchmark dataset}
\label{B.B}
\begin{figure*}[!htp]
	\centering
	{\includegraphics[width = 0.9\textwidth]{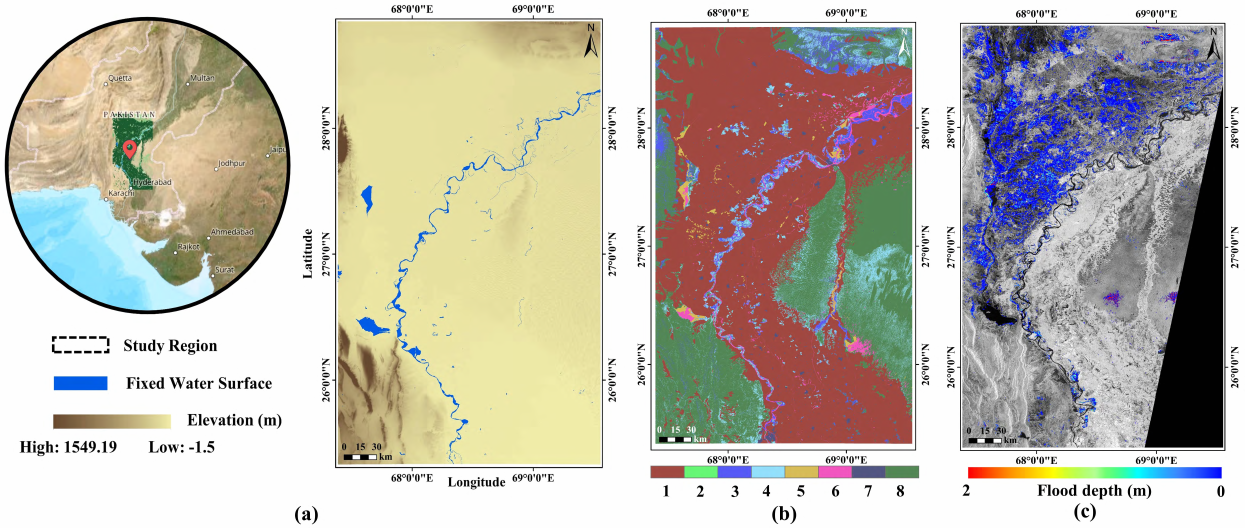}}
	% \vspace{-4mm}
	\caption{Pakistan Flood 2022: (a) Location of the study area and elevation information; (b) land use and land cover. 1: Cultivated land, 2: Forest, 3: Grass land, 4: Shrubland, 5: Wetland, 6: Water body, 7: Artificial surfaces, 8: Bare land; (c) Flood depth is determined using FwDET~\cite{hawker202230}, FABDEM, and SAR-based flood mapping from GloFAS Global Flood Monitoring~\cite{roth2022sentinel} on August 18, 2022.}
	\label{fig:4}
\end{figure*}

\begin{figure*}[!htp]
	\centering
	{\includegraphics[width = 0.8\textwidth]{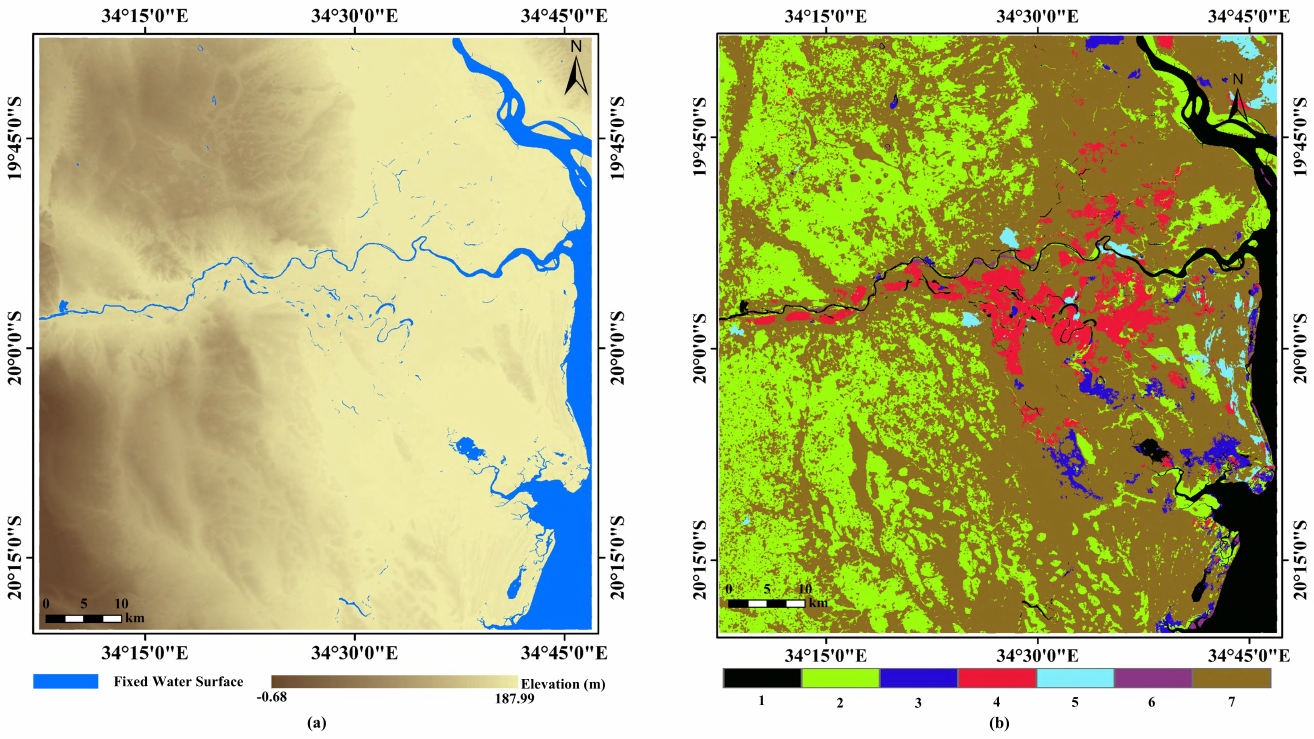}}
	% \vspace{-4mm}
	\caption{Mozambique Flood 2019: (a) Elevation information; (b) land use and land cover. 1: Water, 2: Trees, 3: Flooded vegetation, 4: Crops, 5: Built Area, 6: Bare ground, 7: Rangeland.}
	\label{fig:55}
\end{figure*}

% \begin{figure*}[!htp]
% 	\centering
% 	{\includegraphics[width = 0.85\textwidth]{fig5.pdf}}
% 	% \vspace{-4mm}
% 	\caption{(a) Elevation information of Pakistan Flood 2022; (b) Elevation information of Mozambique Flood 2019}
% 	\label{fig:4}
% \end{figure*}

\subsubsection{Study regions}
\textbf{Pakistan Flood 2022.} Flood events are recurrent phenomena in Pakistan, primarily driven by intense summer monsoon rainfall and occasional tropical cyclones. In the summer monsoon season of 2022, Pakistan experienced a devastating flood event.  This flood event impacted approximately one-third of Pakistan's population, resulting in the displacement of around 32 million individuals and causing the loss of 1,486 lives, including 530 children. 
The economic toll of this disaster has been estimated at exceeding $\$30$ billion~\cite{Bhutto2022}. 
The study area encompasses the regions in Pakistan most severely affected by the flood, spanning the southern provinces of Punjab, Sindh, and Balochistan, covering a total land area of 85,616.5 square kilometers. The Indus River basin, a critical drainage system, plays a pivotal role in this study area's hydrology. Fig.~\ref{fig:4} (a) depicts a visual representation of  Digital Elevation Model (DEM) and river network of Pakistan Flood. A notable change
occurred between August 18 and August 31, 2022, marked
by significant increases in flood coverage within the Pakistan
study area~\cite{xu2024large}.
Consequently, we implement the flood simulation for this
14-day period, totaling 1,209,600 seconds.

\textbf{Mozambique Flood 2019.}
Tropical Cyclone Idai struck the coastal city of Beira, Sofala Province, Central Mozambique on 14 March 2019, unleashing heavy rainfall and powerful winds persisting for over a week. Consequently, the Pungwe and Buzi Rivers, along with lakes, overflowed, inundating low-lying areas~\cite{guo2021mozambique}. This flood caused 4,000 houses to be damaged or inhabitable, about 1,600 people were injured, and 603 people were killed in Mozambique, Zimbabwe, and Malawi~\cite{undp2019}. 
Encompassing the regions within Beira, Mozambique, the study area spans an extensive land area of 6,190.9 square kilometers. Fig.~\ref{fig:55} (a) provides a visual depiction of  DEM and river network of Mozambique Flood. The flood simulation period extends from March 14 to March 20, 2019, according to rainfall records.

\subsubsection{Data requirements for numerical methods}
The data required for conventional numerical methods, such as a finite difference (FD) scheme, encompass topographical data, land cover maps, and real-time gridded rainfall data within the study area.

Specifically, a topographical data, FABDEM~\cite{hawker202230} from COPDEM30, is utilized for flood simulation, with a spatial resolution of 480m $\times$ 480m. 

Land cover information is useful for estimating and adjusting friction (Manning coefficient) in Eq.~\ref{eq44}. Land cover information in Pakistan  can be subtracted from a publicly available Globa-Land30 dataset~\cite{chen2015global} developed by the Ministry of Natural Resources of China, which is shown in Fig.~\ref{fig:4} (b). It is a parcel-based land cover map created by classifying satellite data into 8 classes. Land cover information in Mozambique  can be subtracted from the  Sentinel-2 land use/land cover dataset~\cite{karra2021global}, which is shown in Fig.~\ref{fig:55} (b). It is produced by a deep learning model by classifying Sentinel-2 data into 9 classes. 
Both datasets provide land cover information at a spatial resolution of up to 30 m for their study areas.
% Cultivated land is the predominant land cover type ($57.05\%$) in the Pakistan study area, and urban areas only account for $0.8\%$ of the total study area.
Based on the range suggested by FLO-2D User's Manual~\cite{o1993two} and the land cover types of the study area,  the values of Manning's friction coefficients can be determined. It is worth noting that the infiltration rate in Eq.~\ref{eq44} may not be considered when the infiltration has been basically saturated due to continuous rainfall in the study area.

The rainfall data is a grid-based data set at $0.1^{\circ} \times 0.1^{\circ}$ spatial resolution and half-hourly temporal resolution from GPM-IMERG. Utilizing bilinear interpolation to resample the data, rainfall data with a temporal resolution of 30 seconds and a spatial resolution of 480 m $\times$ 480 m is obtained.

Hydrological stations play a pivotal role in furnishing essential data about inflow boundaries necessary for flood predictions. For Pakistan flood, utilizing the inflow records obtained from a limited number of stations situated along the Indus River, as disclosed by the Government of Pakistan, and through a comparative analysis of the inflow boundaries within the study area against the nearest hydrological station record data, we have computed the daily discharges for inflow boundary from August 18th to August 31st, as presented in Table~\ref{Tab:Table66}. For  Mozambique Flood, due to the absence of a clearly defined inflow boundary, we set the inflow boundary flow to 0.

\begin{table*}[htp!]
	\caption{Daily discharges for inflow boundary from August 18th to August 31th. }
	\centering
	\resizebox{0.9\textwidth}{!}
	{
		\begin{tabular}{cccccccc}
			\hline
			Dates               & 18-Aug & 19-Aug & 20-Aug & 21-Aug & 22-Aug & 23-Aug & 24-Aug \\
			Discharge at inflow ($m^{3}/s$) & 9345   & 9798   & 10251  & 11667  & 13677  & 15914  & 15489  \\ \hline
			Dates               & 25-Aug & 26-Aug & 27-Aug & 28-Aug & 29-Aug & 30-Aug & 31-Aug \\
			Discharge at inflow ($m^{3}/s$) & 14272  & 13875  & 13875  & 13734  & 14187  & 14527  & 14696  \\ \hline
	\end{tabular}}
	\label{Tab:Table66}
\end{table*}

\subsubsection{Implementation details of numerical methods}
We employ a conventional method, as documented in the references~\cite{de2013applicability}, to discretize Eq.~\ref{eq44} from its continuous domain to a discrete domain. This discretization process utilizes the FD scheme applied to a staggered grid. It is worth noting that the FD scheme is a widely accepted numerical solution technique for simulating flood scenarios.

The inputs to a hydraulic simulation include an elevation map, initial conditions,  boundary conditions, and the rainfall conditions in the Pakistan and Mozambique study regions. Specifically, for Pakistan flood 2022, the initial conditions mainly consider the water depths obtained from SAR data (Fig.~\ref{fig:4} (c)). For Mozambique flood 2019, the initial condition is obtained by prerunning FD solver on a dry domain using 1 days of GPM-IMERG rainfall data.
Regarding boundary conditions, for Pakistan flood 2022, our main consideration is the inflow boundary of the Indus River, which spans the study area. The determination of inflow at the upstream extremity of the river is derived from comprehensive hydrological datasets.  We meticulously specify the inflow width as spanning 10 pixels, precisely centered on the lowest point of the river.  For Mozambique Flood, we set the inflow boundary flow to 0.  Rainfall is input as a spatial grid, and Manning coefficients are selected based on land cover and FLO-2D references~\cite{o2011flo}. Furthermore, the discrete implementation described in~\cite{de2013applicability} uses two parameters - a weighting factor $\theta$ that adjusts the amount of artificial diffusion, and a coefficient $0<\alpha \leq 1$ that is used as a factor by which we multiply the time step. We use the proposed values in~\cite{de2013applicability}, namely $\theta=0.7, \alpha=0.7$. 
We implement the numerical solution using Python for a 14-day period (August 18 to August 31), totaling 1,209,600 seconds, in the Pakistan study area (Fig.\ref{fig:4} (a)), and for a 6-day period (March 14 to March 20), totaling 518,400 seconds, in the Mozambique study area (Fig.\ref{fig:55} (a)).

We obtain results at a spatial resolution of 480m $\times$ 480m. Temporal resolution adheres to the Courant-Friedrichs-Lewy (CFL) condition for hyperbolic system stability.
Due to the fine temporal resolution (less than 10s), the 14-day simulations at 480m $\times$ 480m resolution run on an NVIDIA A6000 GPU in about one week. The 6-day simulations on the same GPU take about two days to complete.
\subsubsection{Data generation of flood simulations}
\begin{figure}[!htp]
	% \vspace{-3mm}
	\centering
	{\includegraphics[width = 1.0\textwidth]{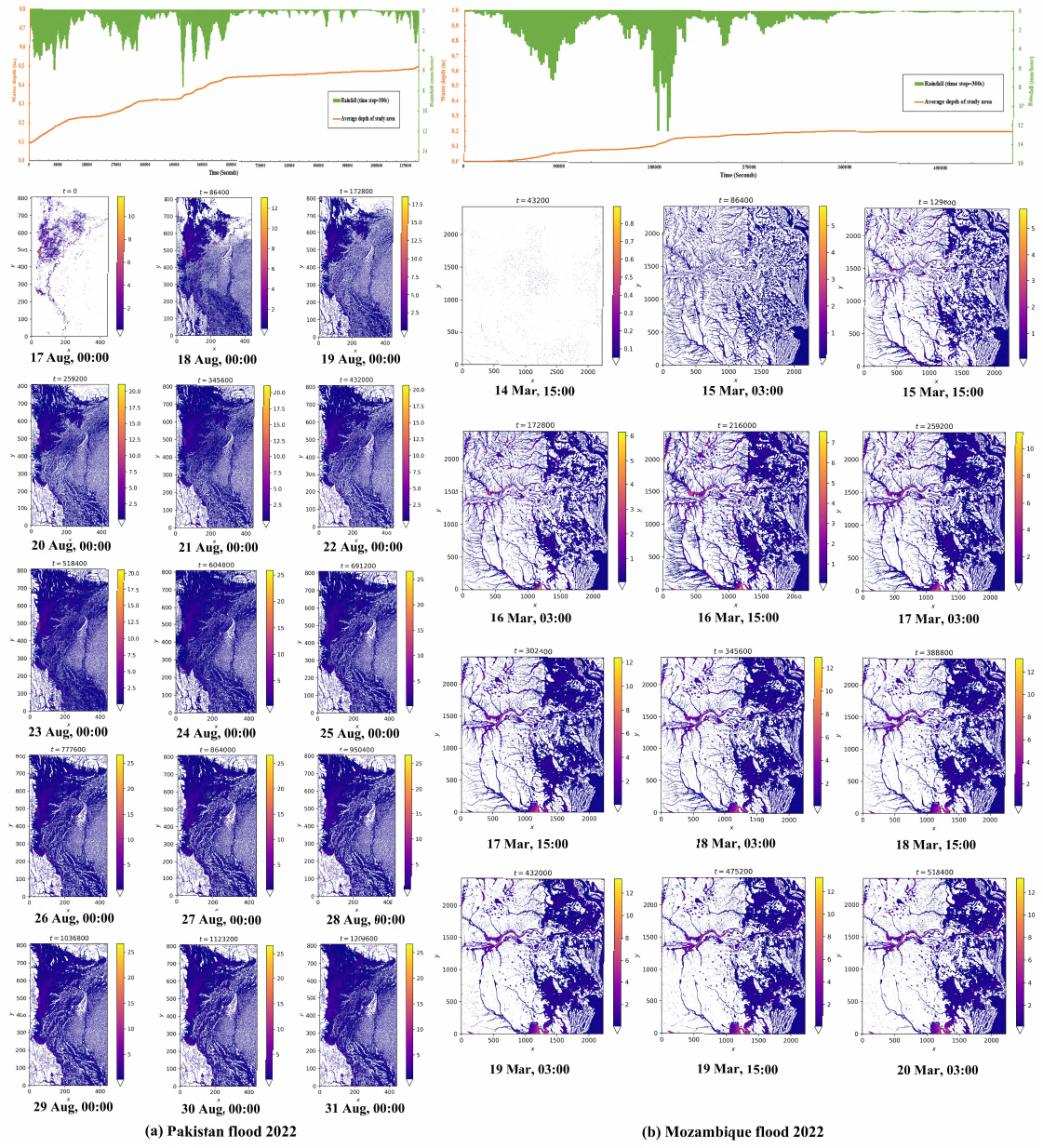}}
	% \vspace{-4mm}
	\caption{(a) Illustration depicting the evolution of the flood simulation in Pakistan in 2022; (b) Illustration depicting the evolution of the flood simulation in Mozambique in 2019. In both figures, the upper curve represents the average depth of the study area, calculated using traditional numerical methods, over a 14-day or 6-day period of rainfall with a spatial resolution of 480m and a time resolution of 30s. The lower portion illustrates the spatial variation of water depth in the study area over the same period.
	}
	\label{fig:3}
\end{figure}

%A notable change occurred between August 18 and August 31, 2022, marked by significant increases in flood coverage within the Pakistan study area~\cite{xu2023large,nanditha2023pakistan}. Consequently, we implement the numerical solution for this 14-day period, totaling 1,209,600 seconds. The spatial resolution is fixed at $480m \times 480m$, and the temporal resolution is 30 seconds for training, validation, and testing. Detailed information regarding data generation for flood simulation, including data requirements and implementation details of numerical methods, is provided in Appendix~\ref{B.B}.
We visualize the evolution of flood heights over a 14-day period in the Pakistan flood and a 6-day period in the Mozambique flood due to rainfall, as shown in Fig.~\ref{fig:3}.

% We set different time scales, $T=20$ and $T=40$, to assess the effectiveness of PeFNN. Specifically, for Pakistan flood, the 14-day sample is  split into three sets: training (1,440 samples for $T=20$, 720 samples for $T=40$), validation (288 samples for $T=20$, 144 samples for $T=40$), and test (288 samples for $T=20$, 144 samples for $T=40$). 
% The 6-day sample from the Mozambique flood is used to verify the transferability of the model trained on the Pakistan flood dataset for cross-regional flood forecasting (432 samples for $T=20$, 864 samples for $T=40$).

\subsection{Training details}
\label{B.C}

\subsubsection{Task details} 
\label{B.C0}
For NS equations, we project the ground truth vorticity field up to $t = 10$ to the field at each time step up to $T > 10$. 
Specifically, for $\nu=1 \times 10^{-3}$, we set $T=50$; for $\nu=1 \times 10^{-4}$, we set $T=30$; for $\nu=1 \times 10^{-5}$, we set $T=20$. 
For SWE, we map the depth of the water at $t = 1$ up to the depth at $T=25$. 
For flood simulations, we set different time scales, $T=20$ and $T=40$, to assess the effectiveness and transferability of models. Specifically, for Pakistan flood, the 14-day sample is  split into three sets: training (1,440 samples for $T=20$, 720 samples for $T=40$), validation (288 samples for $T=20$, 144 samples for $T=40$), and test (288 samples for $T=20$, 144 samples for $T=40$). 
 The 6-day samples with a time step of 30 seconds from the Mozambique flood  are used for cross-regional flood forecasting (432 samples for $T=20$ and 864 samples for $T=40$). For each sample, we map the flood depth at $t = 1$ up to the depth at $T=20$ or $T=40$. 

\subsubsection{Implementation details of PeFNN.} 
\label{B.CA}
Four MC-Fourier layers are utilized with Fourier modes $k$ = 12 in PeFNN, and the channel dimension of the latent space $d_{z}$ = 10. We use Adam optimizer~\cite{kingma2014adam}  with $\beta_1=0.9, \beta_2=0.999$, and weight decay $10^{-4}$. We use 500 epochs for the Recurrent training strategy and 100 epochs for the Markov training strategy, with a cosine learning rate scheduler that starts at 0.001 and is decayed to 0. The model is trained on a single NVIDIA RTX A6000 GPU with 48GB memory.
\subsubsection{Implementation details of benchmarks.} 
\label{B.CB}
We use 4 Fourier layers for all frequency-domain models, including FNO and G-FNO, truncating the transform to the 12 lowest Fourier modes for all models.  For the baseline FNO, the dimension of the latent space is 20. For G-FNO, the dimension of the latent space is 10. 
For non-frequency domain models, the potential feature dimension of the first layer in U-Net is set to 32. In PeRCNN, three layers of $1 \times 1$ convolutional layers in the loop module are employed, with the middle feature dimension set to 8. Additionally, a predefined FD-based filter is utilized as the residual connection in the recurrent module.

Across all models, a batch size of 20 is used for NS equations and a batch size of 8 for SWE and flood simulation. A cosine learning rate scheduler, starting at 0.001
and decaying to 0 is implemented. The Adam optimizer is employed with $\beta_1=0.9, \beta_2=0.999$, and weight decay $10^{-4}$. For the Recurrent and Oneshot training strategies, 500 epochs are used, while the Markov training strategy employs 100 epochs. All models are implemented using PyTorch and trained on a single NVIDIA RTX A6000 48GB GPU.

\subsubsection{Evaluation} 
\label{B.eva}
Relative mean square error $L_{\text{RMSE}}$~\cite{li2020fourier, helwig2023group} between the predicted solutions and the computational fluid dynamics  solutions is regarded as an evaluation measure,
\begin{equation}
L_{\text{RMSE}}=\frac{1}{n} \sum_{i=1}^{n} \frac{\left\|\hat{y}_i-y_i\right\|_2}{\left\|y_i\right\|_2},
\end{equation}
where $\hat{y}_i$ and $y_i$ denote the predicted solution and ground truth of the $i$-th test PDE. $n$ is the number of test PDEs and $\|\cdot\|_2$ is the $L_2$ norm. 

Additionally, we compute the momentum loss $L_{\text{M}}$~\cite{takamoto2022pdebench} to evaluate each method's effectiveness in conserving momentum,
\begin{equation}
L_{\text{M}}=\frac{\left\|\sum M_{\text {pred}}-\sum M_{\text {ref}}\right\|_2}{N},
\end{equation}
 where, $M_{\text{pred}}$ and $M_{\text {ref}}$ represent the predicted and reference momentum, respectively, and $N$ denotes the spatial domain size. 
\section{Additional results and analysis}
\subsection{Robustness of PeFNN}
\label{C.A1}
\begin{figure}[!htp]
	% \vspace{-3mm}
	\centering
	{\includegraphics[width = 0.55\textwidth]{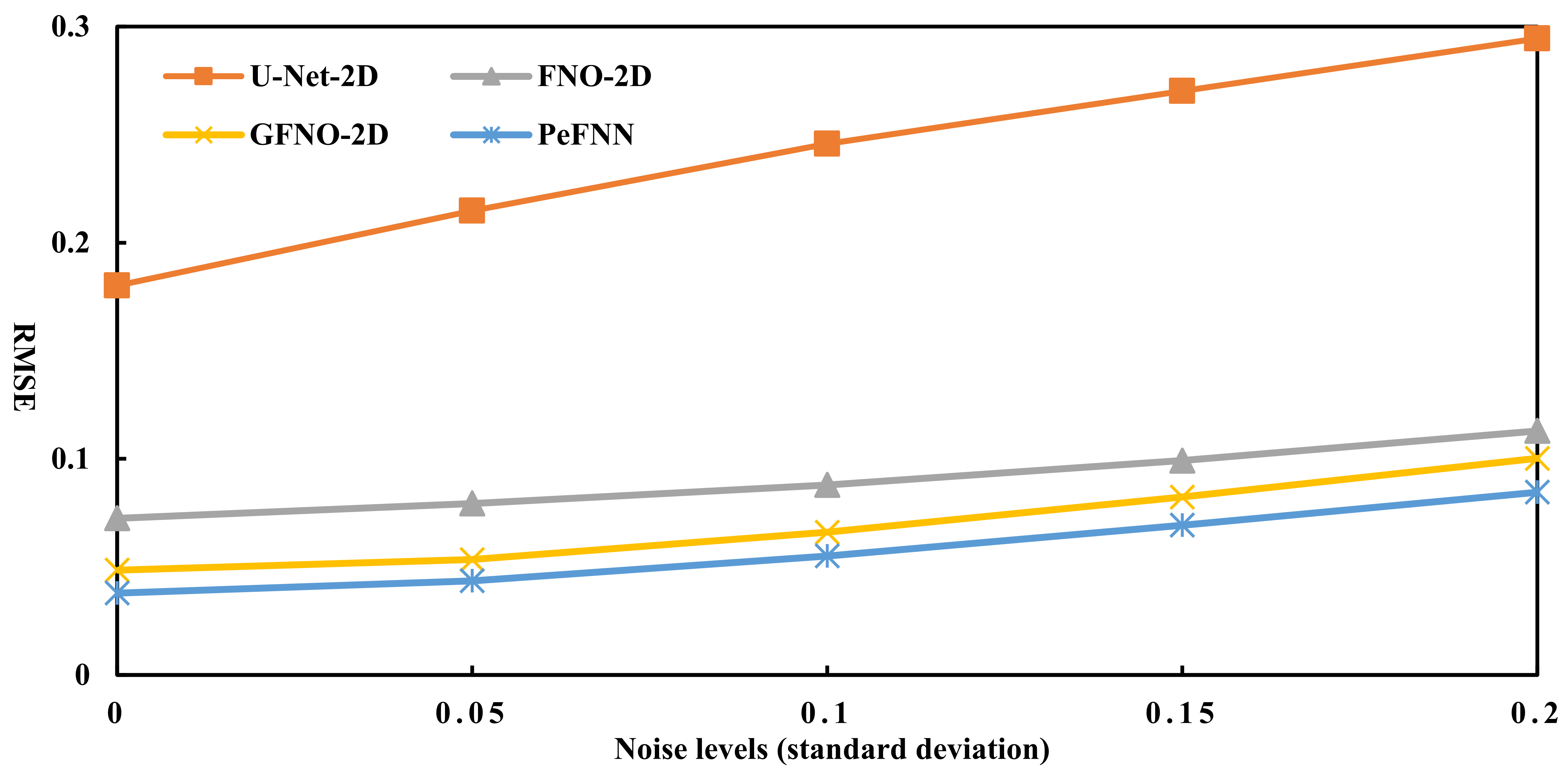}}
	% \vspace{-4mm}
	\caption{ Robustness experiments on the NS equation with $\nu=1 \times 10^{-4}, T=30$. All models are trained using the Markov strategy.
	}
	\label{fig:88}
\end{figure}

We have conducted input data robustness experiments in Fig.~\ref{fig:88}. Specifically, during the model test stage, Gaussian noise of varying levels (standard deviation) is added to the input data to assess the robustness of different models in solving the NS equation.  Notably, our PeFNN consistently outperforms other baselines across different noise levels (standard deviation $\sigma$ = 0, 0.05, 0.1, 0.15, 0.2). Furthermore, even under high levels of Gaussian noise, PeFNN exhibits strong robustness. This demonstrates that PeFNN facilitates automatic generalization and robustness across diverse input data with distributional shifts.

\subsection{SWE super-resolution}
\label{C.A}
\begin{table}[htp!]
	\caption{Super-resolution results on the SWE. We train on a $32 \times 32 \times 24$ grid and directly test on a $128 \times 128 \times 24$ grid, where the first two dimensions are spatial and the third is the number of time steps. Values in parentheses denote standard deviations.}
	\centering
	\resizebox{0.6\textwidth}{!}
	{
		\begin{tabular}{c|cc|cc}
			\hline
			\multirow{2}{*}{Methods} & \multicolumn{2}{c|}{Markov Training} & \multicolumn{2}{c}{Recurrent Training} \\ \cline{2-5} 
			& Test (\%)              & SR Test (\%)          & Test (\%)               & SR Test (\%)           \\ \hline
			U-Net-2D                 & 0.451(0.170)           & 15.732(0.715)          & 3.631(0.514)            & 20.045(4.111)           \\
			FNO-2D                   & 0.180(0.015)           & \textbf{1.515}(0.084) & 0.156(0.009)            & \textbf{1.146}(0.046)  \\
			G-FNO-2D                  &  \underline{0.143}(0.003)     & 2.102(0.428)          & \underline{0.139}(0.058)      &  \underline{1.149}(0.009)     \\
			PeFNN                    & \textbf{0.139}(0.018)  &  \underline{1.614}(0.104)    & \textbf{0.114}(0.038)   & 2.951(0.163)           \\ \hline
	\end{tabular}}
	\label{Tab:Table6}
\end{table}
In Table~\ref{Tab:Table6}, we present super-resolution results for SWE. PeFNN exhibits the lowest test error among all methods. However, the super-resolution test error of PeFNN is marginally lower than FNO-2D. Notably, the super-resolution performance of PeFNN will decrease slightly when the training sample size is reduced. In addition, the super-resolution performance of PeFNN is better than G-FNO-2D  in the Markov Training strategy.

\subsection{Consecutive flood prediction results}
\label{C.B}
\begin{figure}[!tph]
	\centering
	{\includegraphics[width = 1.0\textwidth]{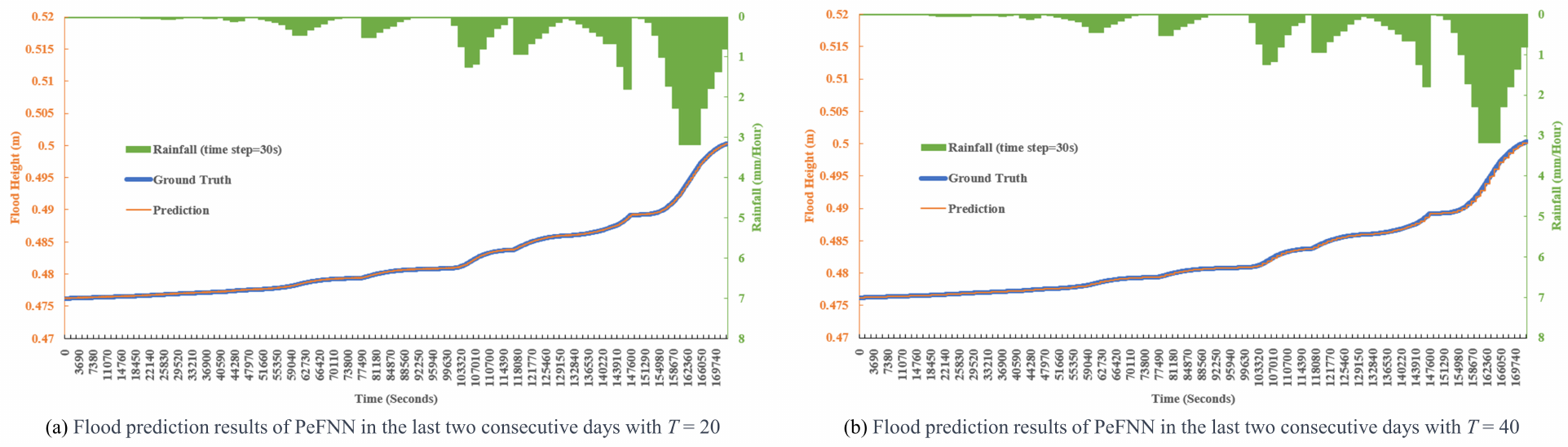}}
%	\vspace{-4mm}
	\caption{Prediction results in Pakistan flood 2022 for two consecutive days with $T=20$ and $T=40$.
	}
	\label{fig:12}
\end{figure}

\begin{figure}[!tph]
	\centering
	{\includegraphics[width = 1.0\textwidth]{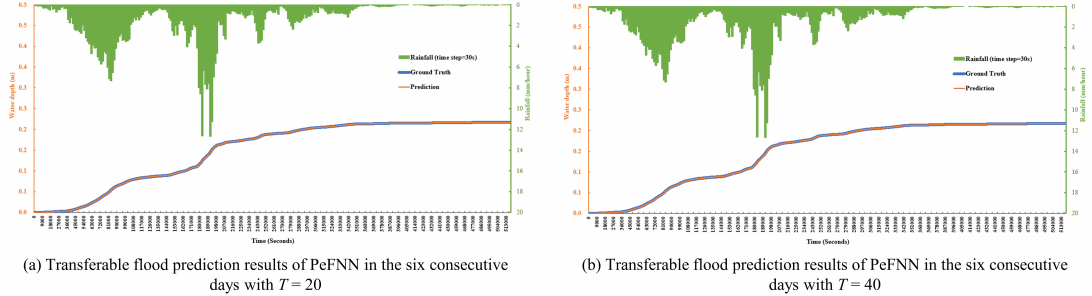}}
%	\vspace{-4mm}
	\caption{Transferable prediction results in Mozambique flood 2019 for six consecutive days with $T=20$ and $T=40$.
	}
	\label{fig:13}
\end{figure}
For Pakistan flood 2022, in our forecasts for the last two consecutive days (August 30th and August 31st), spanning a total of 172,800 seconds, we considered two models with different time scales $T=20$ and $T=40$. We compared the average predicted flood heights of the study area with the actual average heights. The results are illustrated in Fig.~\ref{fig:12} (a) and Fig.~\ref{fig:12} (b). Notably, both prediction results closely match the flood heights of ground truth. Upon further comparison between the two different time scales, we observed that the model trained on the longer time scale ($T=40$) exhibited slightly inferior prediction results in the last period compared to the model trained on the shorter time scale ($T=20$). This highlights the importance of selecting an appropriate time scale for real flood prediction. 

Furthermore, we directly utilize the PeFNN trained on the Pakistan flood dataset to predict the flood dynamics in Mozambique over 6 consecutive days (March 14 to March 20), totaling 518,400 seconds. We evaluate  the transferability of PeFNN  with different time scales $T=20$ and $T=40$. We compare the average predicted flood heights in the Mozambique study area with the actual average heights, as depicted in Fig. \ref{fig:13} (a) and Fig. \ref{fig:13} (b). Both prediction results closely match the actual flood heights, demonstrating that PeFNN can achieve accurate cross-scenario flood predictions at different time scales.

\section{Rollout visualizations}
\label{D}
\begin{figure}[!tph]
	\centering
	{\includegraphics[width = 1.0\textwidth]{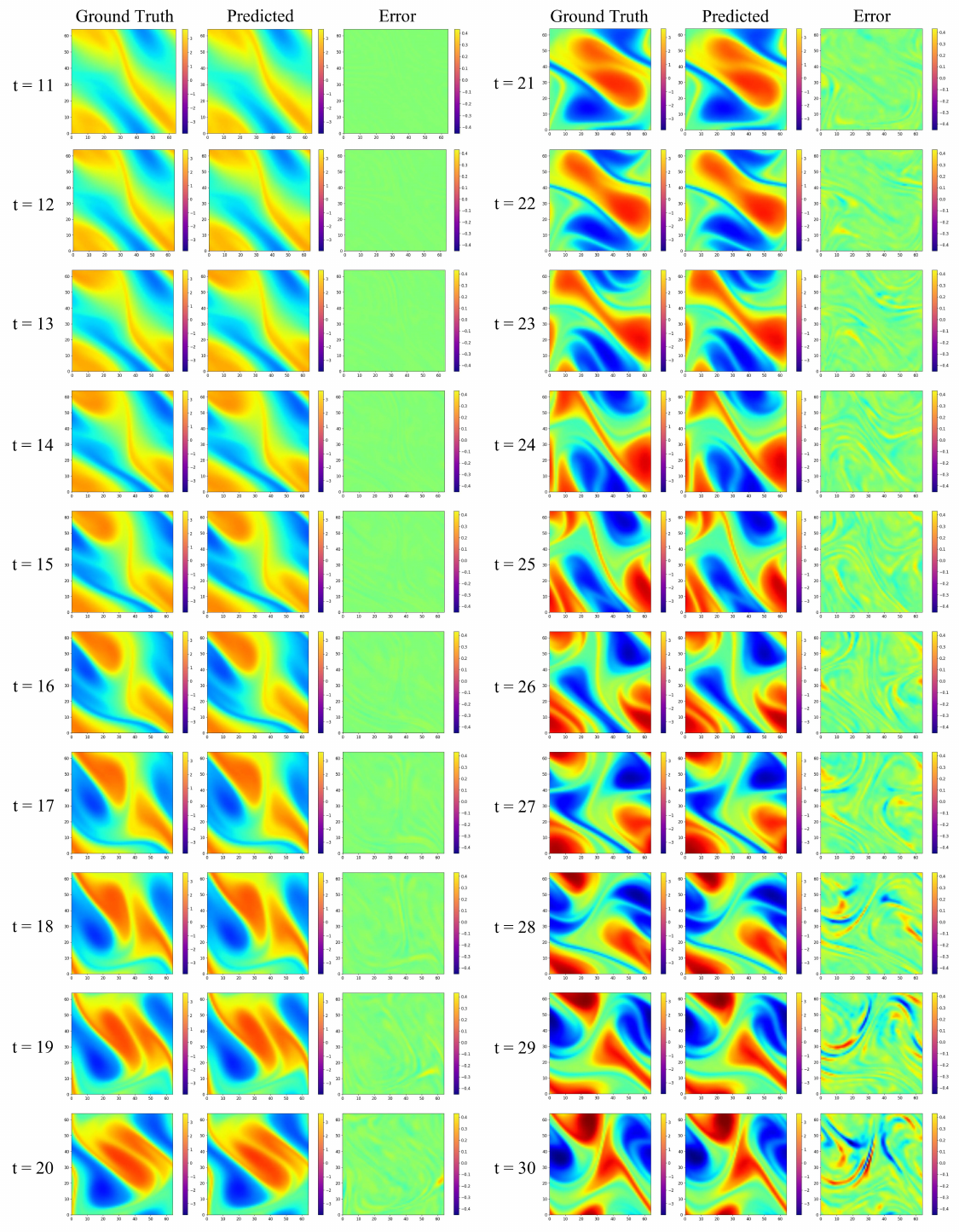}}
%	\vspace{-4mm}
	\caption{Rollout predictions of PeFNN on NS equations under the \textit{Markov} training strategy is conducted by projecting the ground truth vorticity field up to $t = 10$ to the field at each time step up to $t=30$. The resolution is fixed at $64 \times 64$.
	}
	\label{fig:6}
\end{figure}

\begin{figure}[!tph]
	\centering
	{\includegraphics[width = 1.0\textwidth]{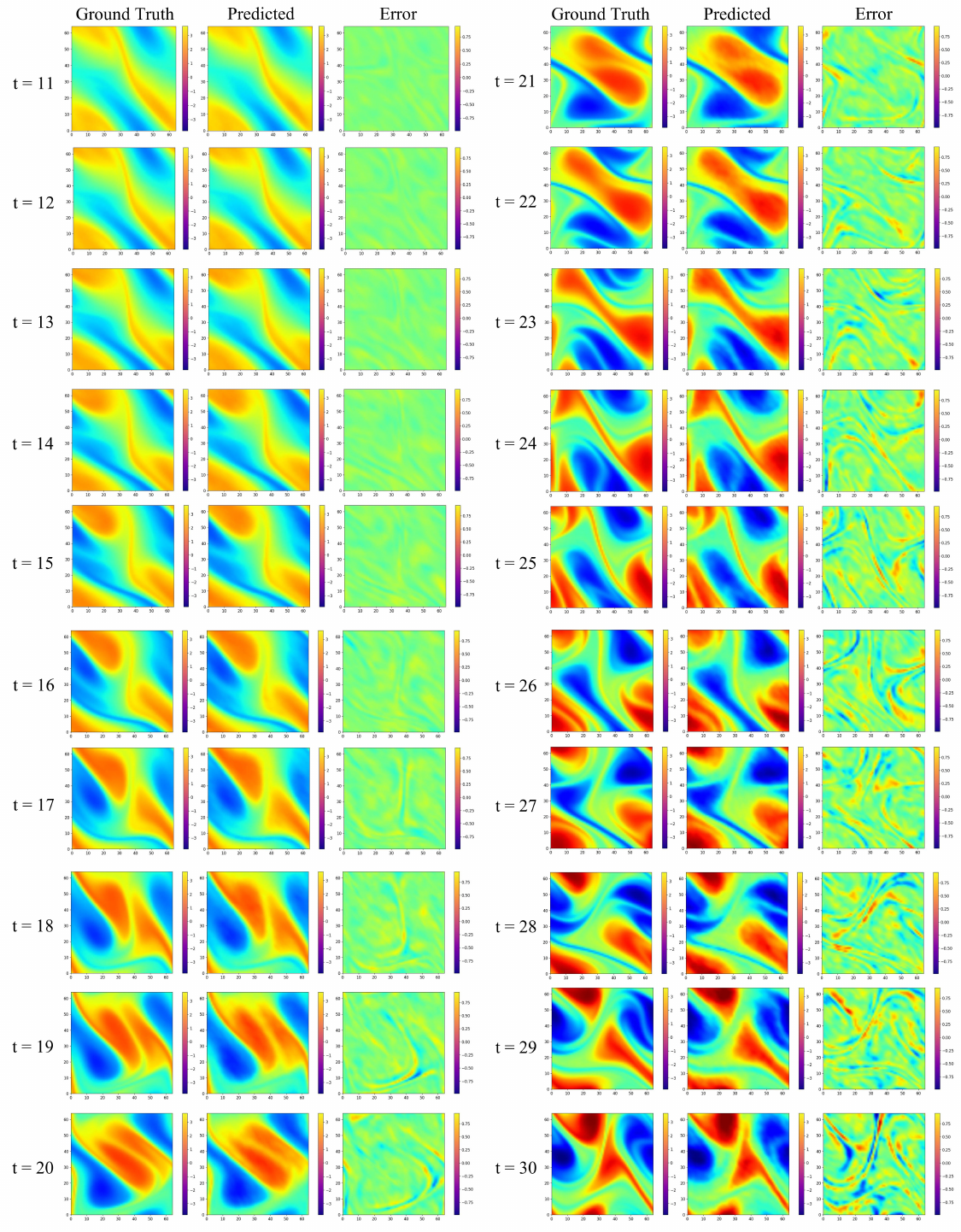}}
%	\vspace{-4mm}
	\caption{Rollout predictions of PeFNN on NS equations under the \textit{Recurrent} training strategy is conducted by projecting the ground truth vorticity field up to $t = 10$ to the field at each time step up to $t=30$. The resolution is fixed at $64 \times 64$.
	}
	\label{fig:7}
\end{figure}

\begin{figure}[!tph]
	\centering
	{\includegraphics[width = 0.85\textwidth]{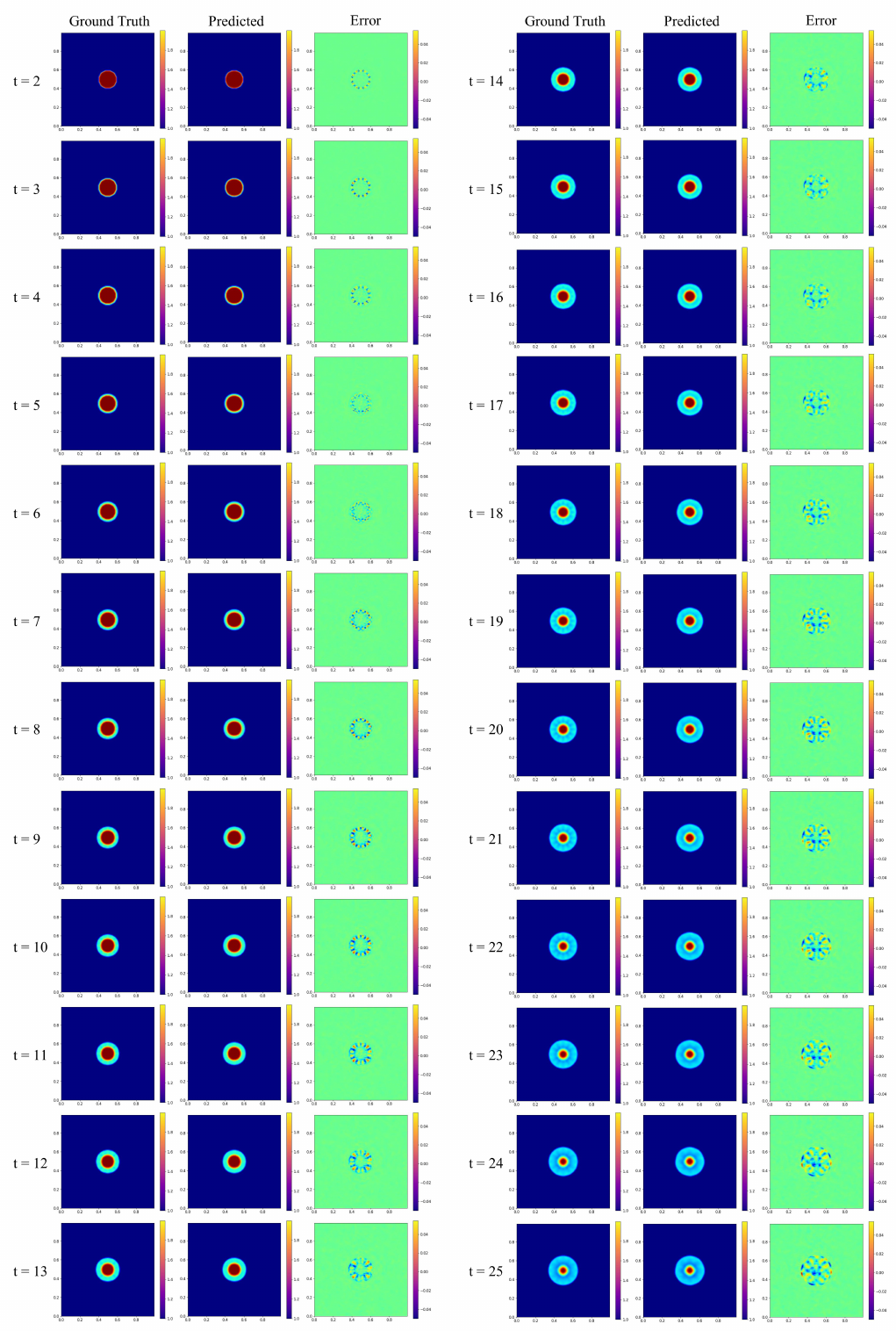}}
%	\vspace{-4mm}
	\caption{Rollout predictions of PeFNN on SWE under the \textit{Markov} training strategy is conducted by mapping the depth of the water at $t = 1$ up to the depth at $t=25$. The resolution is fixed at $128 \times 128$.
	}
	\label{fig:8}
\end{figure}

\begin{figure}[!tph]
	\centering
	{\includegraphics[width = 0.85\textwidth]{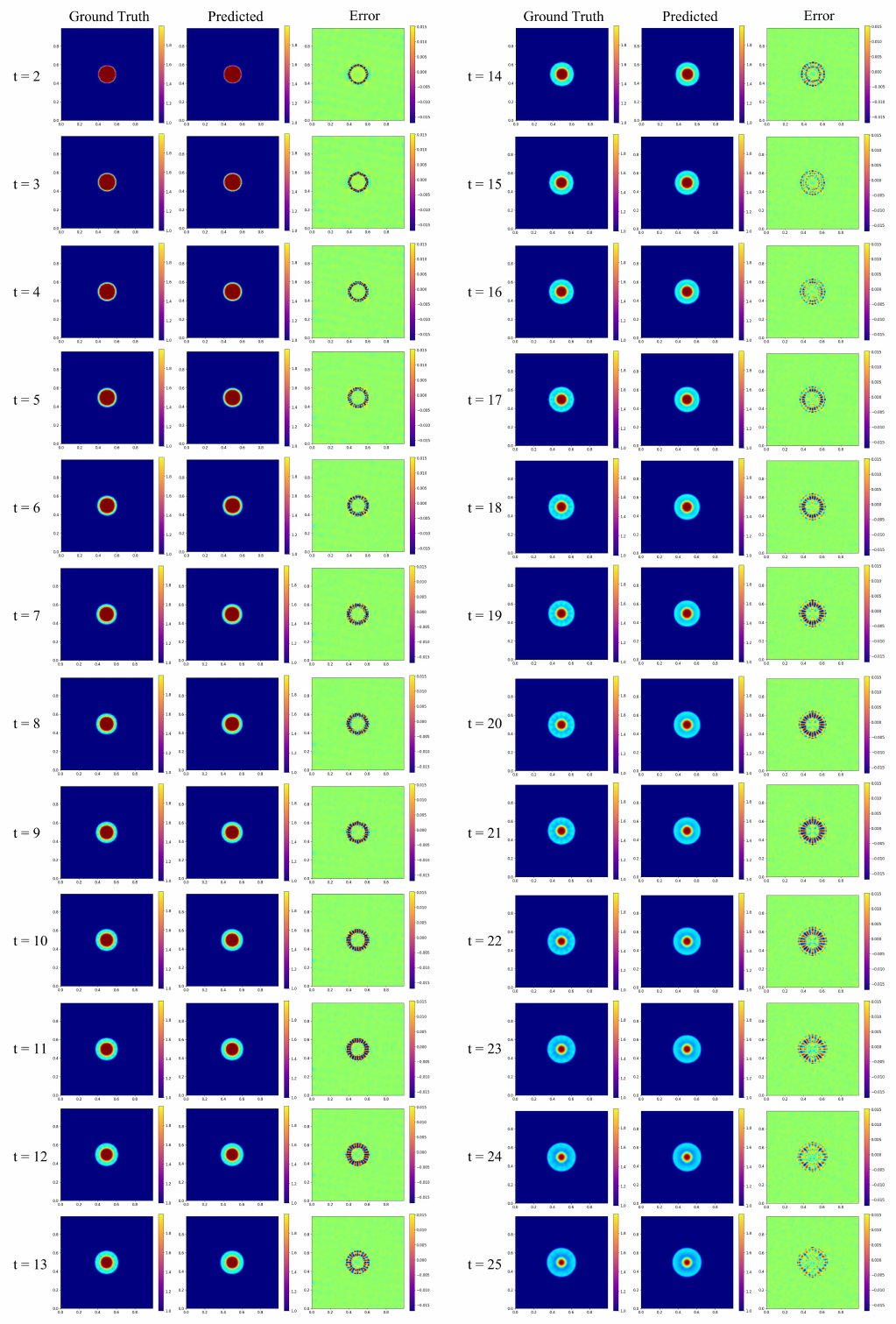}}
%	\vspace{-4mm}
	\caption{Rollout predictions of PeFNN on SWE under the \textit{Recurrent} training strategy is conducted by mapping the depth of the water at $t = 1$ up to the depth at $t=25$. The resolution is fixed at $128 \times 128$.
	}
	\label{fig:9}
\end{figure}

\begin{figure}[!tph]
	\centering
	{\includegraphics[width = 0.85\textwidth]{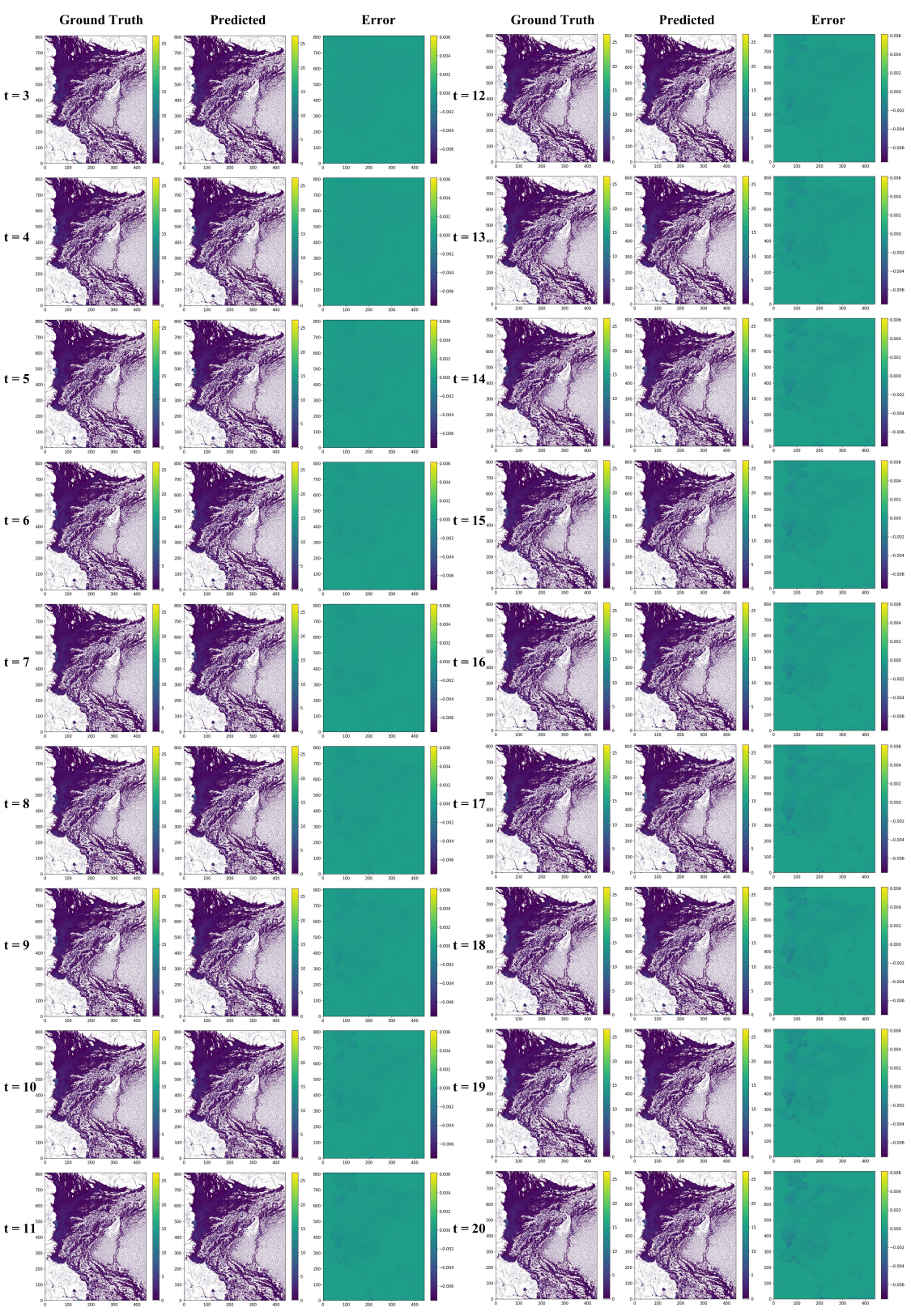}}
%	\vspace{-4mm}
	\caption{Rollout predictions of PeFNN on Pakistan flood simulation under the \textit{Markov} training strategy is conducted by mapping the flood depth at $t = 1$ up to the depth at $t=20$. 
	}
	\label{fig:10}
\end{figure}

\begin{figure}[!tph]
	\centering
	{\includegraphics[width = 1.0\textwidth]{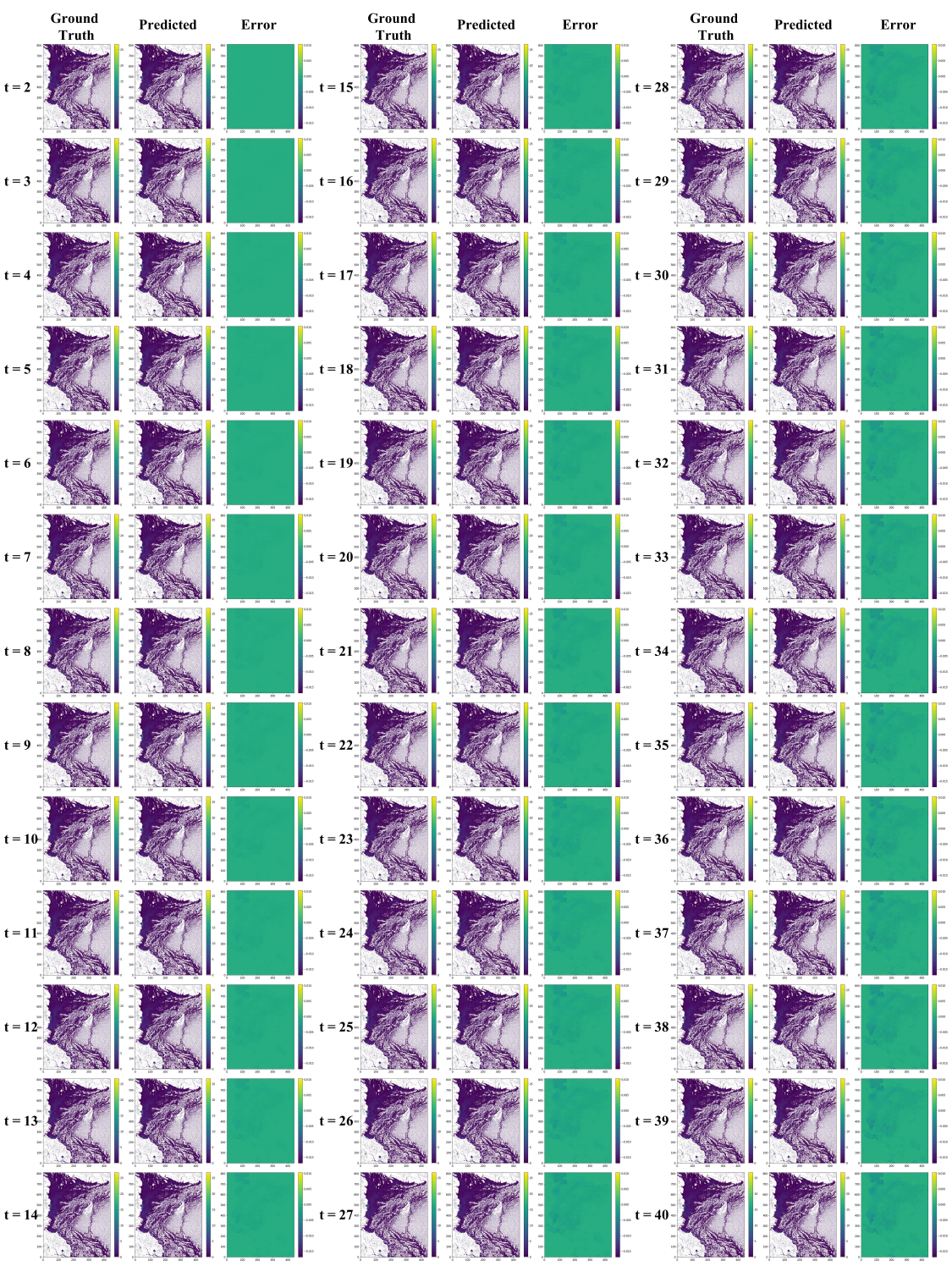}}
%	\vspace{-4mm}
	\caption{Rollout predictions of PeFNN on Pakistan flood simulation under the \textit{Markov} training strategy is conducted by mapping the flood depth at $t = 1$ up to the depth at $t=40$. 
	}
	\label{fig:11}
\end{figure}

\begin{figure}[!tph]
	\centering
	{\includegraphics[width = 0.85\textwidth]{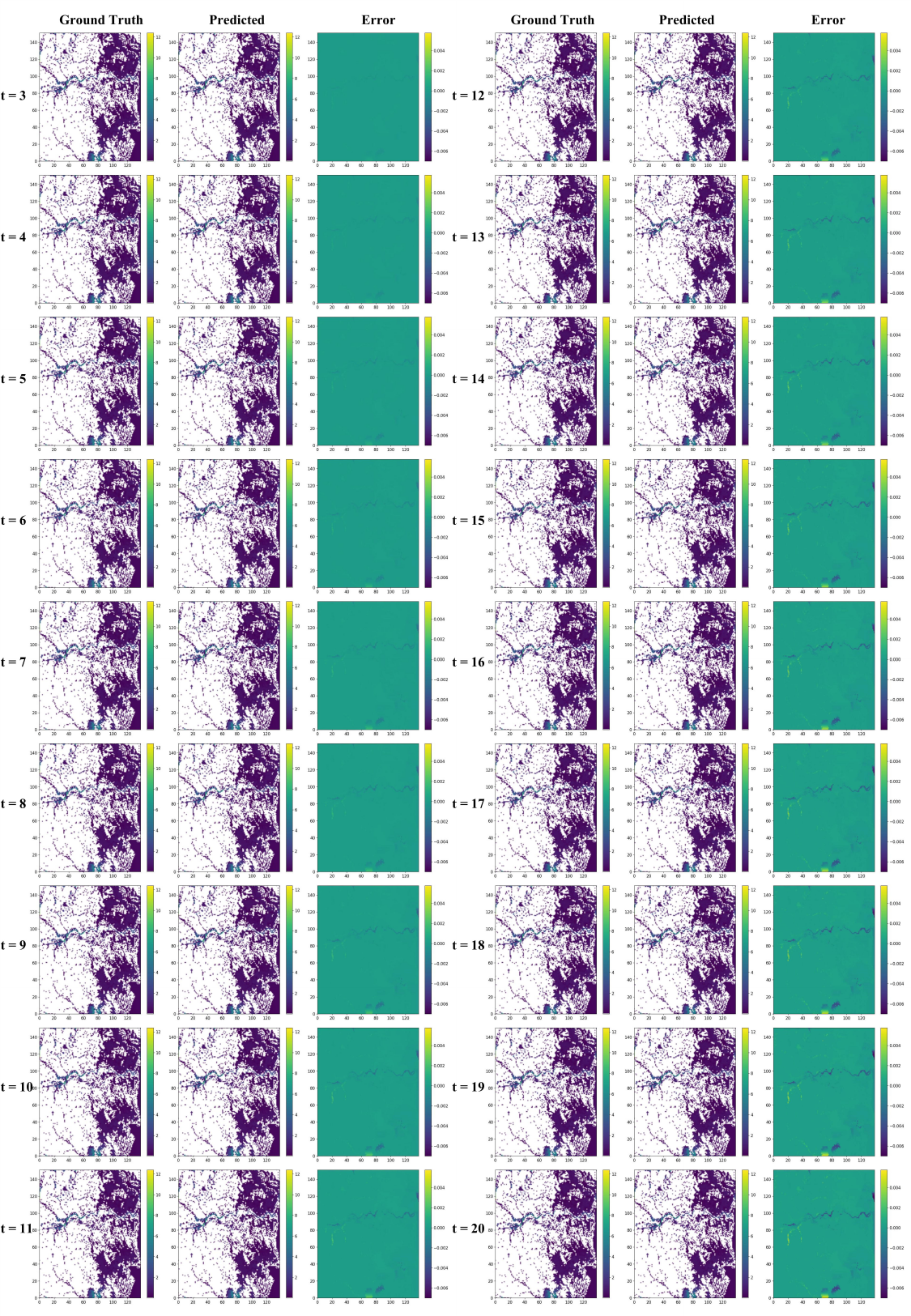}}
	%	\vspace{-4mm}
	\caption{Rollout transferable predictions of PeFNN on Mozambique flood simulation under the \textit{Markov} training strategy is conducted by mapping the flood depth at $t = 1$ up to the depth at $t=20$. 
	}
	\label{fig:101}
\end{figure}

\begin{figure}[!tph]
	\centering
	{\includegraphics[width = 1.0\textwidth]{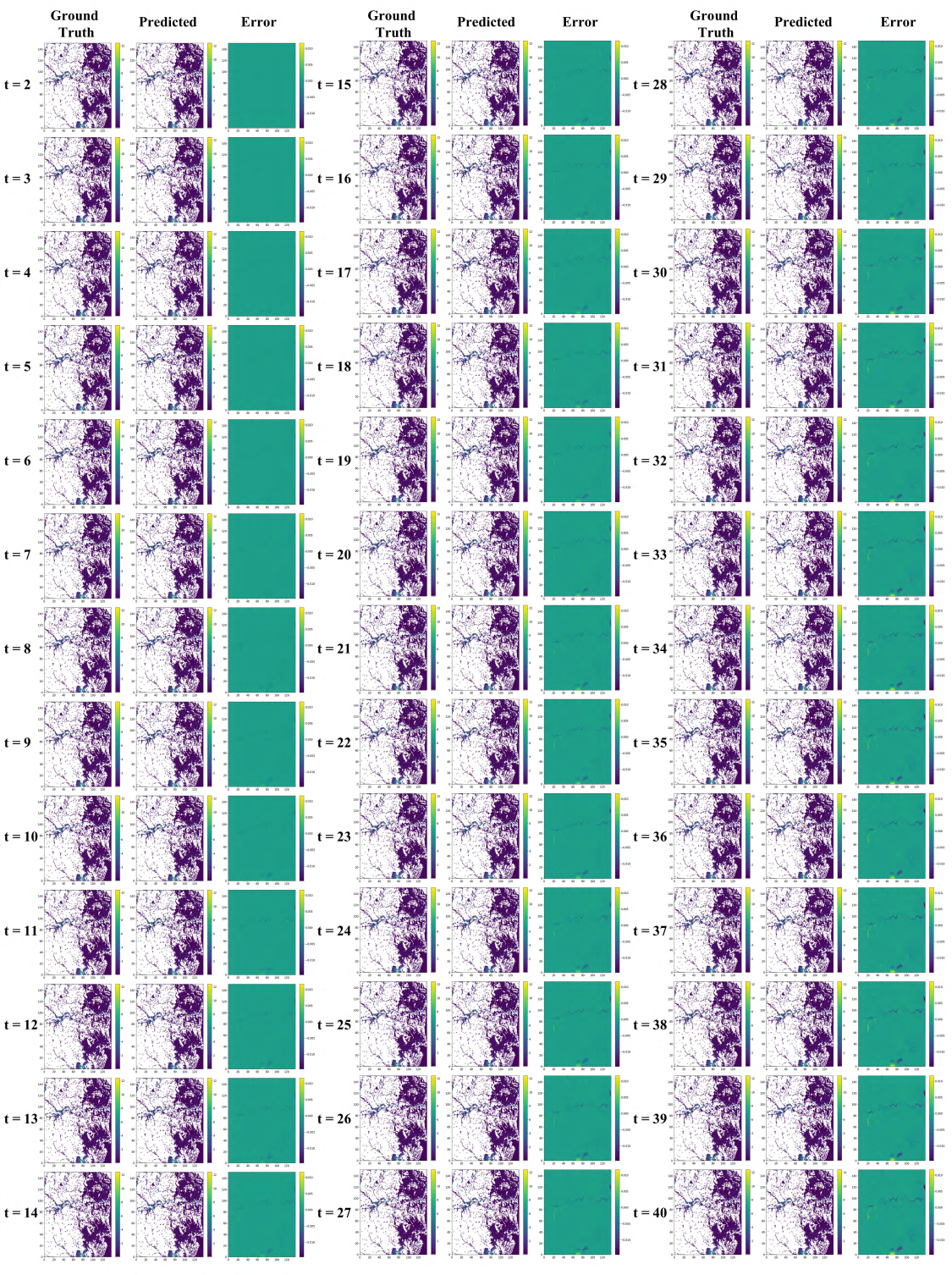}}
	%	\vspace{-4mm}
	\caption{Rollout transferable predictions of PeFNN on Mozambique flood simulation under the \textit{Markov} training strategy is conducted by mapping the flood depth at $t = 1$ up to the depth at $t=40$. 
	}
	\label{fig:111}
\end{figure}
\newpage

\end{document}